\documentclass[10pt,twocolumn,letterpaper]{article}

\usepackage[pagenumbers]{cvpr} %

\usepackage{url}
\usepackage{placeins}
\usepackage{graphicx}
\usepackage{booktabs}
\usepackage{siunitx}
\usepackage{subcaption}
\usepackage{multirow}
\usepackage{amsthm}

\definecolor{cvprblue}{rgb}{0.21,0.49,0.74}
\usepackage[pagebackref,breaklinks,colorlinks,allcolors=cvprblue]{hyperref}

\def\paperID{22066} %
\def\confName{CVPR}
\def\confYear{2026}

\renewcommand{\thefootnote}{\fnsymbol{footnote}}

\title{Your Latent Mask is Wrong: Pixel-Equivalent Latent Compositing for Diffusion Models}

\author{Rowan Bradbury\thanks{Equal contribution.},\quad Dazhi Zhong\footnotemark[1]\\
Bradbury Group\\
New York City, USA\\
\texttt{\{rowan,dazhi\}@bradburygroup.org} \\
}

\begin{document}
\maketitle

\setcounter{footnote}{0}
\renewcommand{\thefootnote}{\arabic{footnote}}

\begin{abstract}
Latent inpainting in diffusion models still relies almost universally on linearly interpolating VAE latents under a downsampled mask. We propose a key principle for compositing image latents: Pixel-Equivalent Latent Compositing (PELC). An equivalent latent compositor should be the same as compositing in pixel space. This principle enables full-resolution mask control and true soft-edge alpha compositing, even though VAEs compress images 8x spatially. Modern VAEs capture global context beyond patch-aligned local structure, so linear latent blending cannot be pixel-equivalent: it produces large artifacts at mask seams and global degradation and color shifts. 

We introduce DecFormer, a 7.7M-parameter transformer that predicts per-channel blend weights and an off-manifold residual correction to realize mask-consistent latent fusion. DecFormer is trained so that decoding after fusion matches pixel-space alpha compositing, is plug-compatible with existing diffusion pipelines, requires no backbone finetuning and adds only 0.07\% of FLUX.1-Dev’s parameters and 3.5\% FLOP overhead.

On the FLUX.1 family, DecFormer restores global color consistency, soft-mask support, sharp boundaries, and high-fidelity masking, reducing error metrics around edges by up to 53\% over standard mask interpolation. Used as an inpainting prior, a lightweight LoRA on FLUX.1-Dev with DecFormer achieves fidelity comparable to FLUX.1-Fill, a fully finetuned inpainting model. While we focus on inpainting, PELC is a general recipe for pixel-equivalent latent editing, as we demonstrate on a complex color-correction task.
\end{abstract}

\section{Introduction}

Latent diffusion models (LDMs)~\citep{rombach2022high,Peebles2022DiT} dominate modern image generation, yet a brittle operation is widespread for mixing image latents. In masked-conditioned generation tasks such as inpainting or editing, latents are interpolated via a mask the same way as pixels. However, this heuristic is a source of error which limit mask fidelity. VAE decoders are nonlinear and spatially entangled, so mixing latents does not mix pixels. The result is off-manifold seams, color shifts, and halos that diffusion then amplifies across denoising steps.

We propose a simple principle: latent compositing should be \emph{pixel-equivalent} (PE). For a frozen encoder $E$, decoder $D$, and any pixel-space operator $F$, a latent operator $C_F$ should satisfy
\begin{equation}
D\!\left(C_F(z)\right)\;=\;F\!\left(D(z)\right),
\hspace{0.5em} 
C_F(E(x)) \; = \; E(F(x))
\label{eq:def_DE}
\end{equation}
That is, applying $F$ after decoding should match applying $C_F$ before decoding, and the same for encoding. We call these two equalities decoder equivalence (DE) and encoder equivalence (EE). As a concrete case, inpainting uses

\begin{equation}
F(x_A,x_B,M)=(1-M)\odot x_A + M\odot x_B
\label{eq:latent_composition}
\end{equation}
To satisfy DE and EE, it must uphold
\[
\begin{aligned}
D\!\left(C_F(z_A,z_B,M)\right)= (1-M)\odot D(z_A)+M\odot D(z_B) \\
\text{and} \hspace{0.5em} C_F(E(z_A),E(z_B),M)= E((1-M)\odot x_A+M\odot x_B)
\end{aligned}
\]

\begin{figure*}
    \centering
    \includegraphics[width=\textwidth]{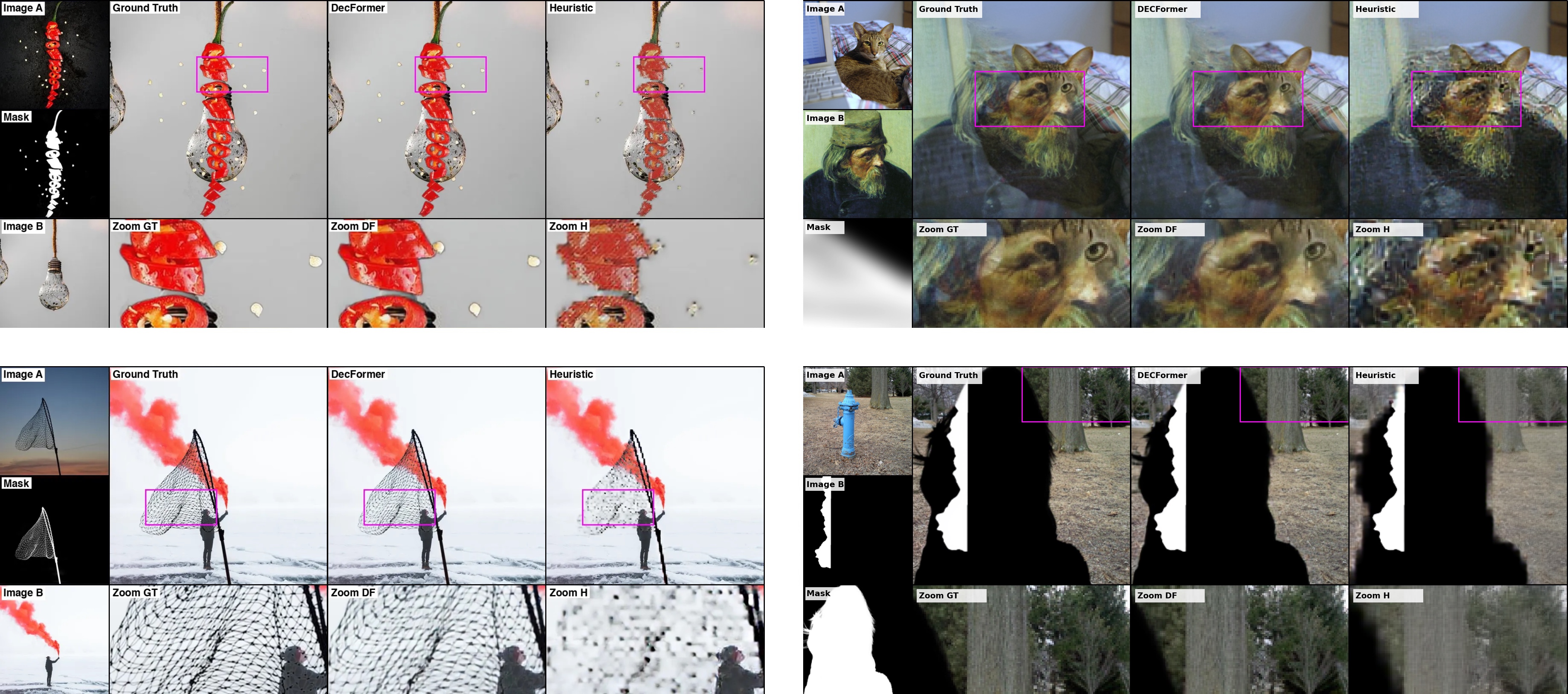}
    \caption{Each quadrant compares ground-truth pixel composites\protect\footnotemark, our DecFormer predictions, and heuristic latent interpolation. Across soft, binary, and structured masks, DecFormer restores sharp edges and high-frequency detail, whereas the heuristic exhibits smearing and artifacts on soft blends, halos and discoloration at boundaries, and blocky low-fidelity masks. Notably, in the bottom-right example, global background degradation occurs far from the masked region, reflecting how latent entanglement corrupts off-mask content; this effect is eliminated by DecFormer.}
    \label{fig:hero}

\end{figure*}

Linear latent blending would satisfy decoder-consistency only if $D$ were locally linear and channel-separable, assumptions that we empirically show fail in modern VAEs (Table~\ref{tab:decformer-interp-metrics}, Fig.~\ref{fig:hero}).

Modern VAE latents couple wide spatial context and heterogeneous channels; broadcasting a single, downsampled mask and linearly mixing latents introduces boundary leakage and global color drift. Figure~\ref{fig:hero} shows the effect on a latent mixing task: heuristic blending yields visible halos and boundary mismatch, while a pixel-equivalent compositor restores sharp edges and global image quality even away from the edge seams. 

\footnotetext{Validation images and alpha mattes not included in training set; sourced from withoutBG and COCO 2017.}

We introduce \textbf{PELC} (Pixel-Equivalent Latent Compositing), a model-agnostic methodology for learning latent operators $C_\theta$ that are decode-equivalent with target pixel operators using only a frozen encoder–decoder and synthetic supervision from pixel composites. As an inpainting instantiation, we propose \textbf{DecFormer}, a lightweight 7.7M-parameter transformer that predicts per-channel blend weights together with a nonlinear residual correction, supporting \emph{genuinely soft masks}, and is a drop-in replacement for heuristic latent compositing. The same principle of pixel-equivalence apply to non-compositing operations as well, such as the color corrections demonstrated in this work.

\paragraph{Contributions}
\begin{itemize}
\item  We formalize pixel-equivalence as a general criterion for latent operators and present PELC, a simple training recipe to realize latent compositing from pixel-space supervision.
\item We formalize and demonstrate (Fig.~\ref{fig:hero}, Table~\ref{tab:decformer-interp-metrics}) that linear latent interpolation cannot meet pixel-equivalence in modern VAEs, which exhibit nonlinearity and wide effective receptive fields.
\item We design DecFormer, a 7.7M-parameter compositor that restores mask fidelity and supports genuinely soft masks with negligible overhead (3.5\% FLOP overhead). Decformer consistently halves error metrics at key mask edge areas.
\item For inpainting, Decformer improves all visual metrics as a drop-in replacement, and adding a lightweight LoRA achieves comparable visual quality with a dedicated inpainting model (Flux-Fill).
\item Additionally to inpainting, our PE objective applies to any pixel-space operator $F$, providing a path to principled latent-space editing without repeated encoding and decoding at every step.
\end{itemize}

\section{Background}
\subsection{LDMs and VAE latents}

Instead of denoising in pixel space, modern diffusion models operate in the latent space of a pretrained variational autoencoder (VAE). We investigate Flux's VAE \citep{bfl2025flux1kontextflowmatching}, latest and state-of-the-art in the line of autoencoders following \cite{rombach2022high}.
Given an image $x \in \mathbb{R}^{H\times W \times 3}$, the VAE encoder $E$ produces a latent tensor
$$
z = E(x) \in \mathbb{R}^{h\times w \times C}, \quad h=H/f,w=W/f
$$
where $f$ is the downsampling factor of the VAE. The decoder $D$ then reconstructs pixels through $x \approx \hat{x} = D(z) $. 

A notable feature of these latents is that they resemble images. Channel-wise visualizations, show downsampled content aligned to the spatial $(h,w)$ grid. The convolution-based architecture creates inductive bias for latents to be spatially consistent with the encoded image. Each latent voxel $z[i,j,:]$ is aggregated by strided convolutions over a receptive field centered at approximately
$$
\Bigl( i \cdot f ,\;  j \cdot f \Bigr).
$$
The effective stride $S_E = \prod_{i=l+1}^{L}s_i$ cumulatively over encoder layers $L$ and their respective strides $s_i$ is equal to the downsample factor $f$. In Flux's autoencoder, $S_l=f=8$. Because convolutions are translation-equivariant, and each latent position's receptive field over the pixel space is strided by $f$, latents features of the autoencoder are exactly spatially correlated with the image by patches of $f\times f$.

\paragraph{Latent Diffusion}  Flux is a rectified flow latent diffusion model (LDM) predicting a mapping between a noise distribution $\epsilon  \sim \mathcal{N}(0,1)$ to latent samples $z$ through a velocity field $v_\theta(z_t,t)$,  defined as an ordinary differential equation
$$
dz_t = v_\theta(z_t, t) dt
$$
Thus, we have flow sampling process given $t\in [0,1]$
\[
z_{t'} \;\leftarrow\; z_t + (t' - t)\, v_\theta(z_t, t).
\]

\subsection{Heuristic latent masking} \label{sec:heuristic_masking_bg}
This image-like structure motivated the heuristic of applying masks directly in the latent space for inpainting, which is now common practice in commercial use, academic literature \cite{couairon2023diffedit} and mainstream pipelines. Concretely a pixel mask, often binary but common with softened edges \footnote{Common inpainting pipelines, including Diffusers, apply edge-softening operations such as blurring or morphological smoothing when preparing masks.}, $M\in\{0,1\}^{H\times W\times1}$, is interpolated to match the latent resolution of $z$ with a fixed downsampler $\mathcal{S}:\{0,1\}^{H\times W}\to[0,1]^{h\times w}$ and the result latent mask $m$ is broadcast to match the channel dimensions of $z$.

\[
\begin{aligned}
m \;&=\; \mathcal{S}(M)\in[0,1]^{h\times w}, \\
m[i,j,:]\;&=\; m[i,j]\,\mathbf{1}_C.
\end{aligned}
\]
We write $\alpha$ for a full-channel latent mask; we broadcast $m$ across all latent channels for heuristic masking.
At each sampling step, the latent update $\hat{z}_{t\rightarrow t-1}$ is convex blended with the original latent $z_{t-1}^{orig}$ using latent mask $m$.
\begin{equation}
z_{t-1} \;=\; (1-m)\odot \hat z_{t\to t-1} \;+\; {m}\odot z^{\text{orig}}_{t-1}.
\label{eq:heuristic_masking_latent}
\end{equation}
Throughout our paper we refer to this approach as heuristic blending. This approach of blending latents like images worked \textit{well enough} in early VAEs such as Stable Diffusion (SD) 1.x/2.x/XL, whose modest 4-channel latents were relatively image-like and whose decoder receptive fields were relatively narrow. But the justification is mathematically unsound. The autoencoder does \textit{not} satisfy
$$
E(x_A \oplus_M x_B) \;=\; (1-m) \cdot E(x_A) + m\cdot E(x_B),
$$

so convex mixing in latent space is not guaranteed to correspond to masked mixing in pixel space.

\paragraph{Observation}
\label{prop:impossibility}
For a nonlinear VAE decoder $D$, there exist $z_A,z_B$ and masks $M$ such that
\[
D((1-\alpha)z_A+\alpha z_B) \neq (1-M)\odot D(z_A)+M\odot D(z_B)
\]
for all $\alpha\in[0,1]^{h\times w\times C}$. That is, no convex latent interpolation is pixel-equivalent in general—a consequence of decoder nonlinearity that we demonstrate empirically in modern VAEs.

\subsection{Heuristic masking pathology} \label{sec:heuristic_pathology}

\begin{figure}[t]
    \centering
    \includegraphics[width=\linewidth]{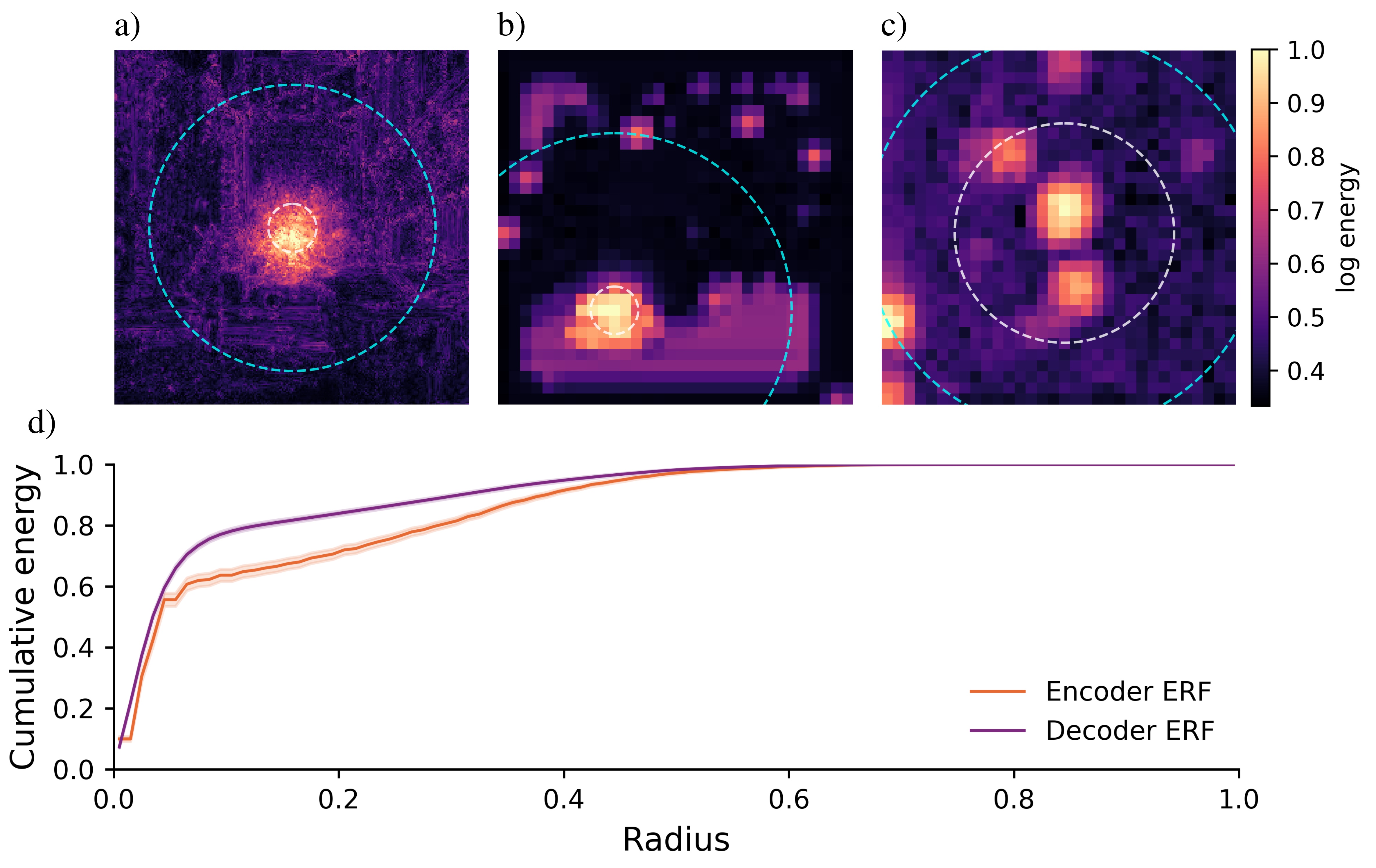}
    \caption{\textbf{Effective Receptive Field (ERF) analysis} of the Flux VAE on $256{\times}256$ images; panels (a–c) are randomly chosen near the pooled median. Statistics pool $128{\times}3$ probes with 95\% bootstrap CIs; radii $r_{50},r_{90}$ are cumulative-energy radii (fraction of image/latent diagonal). \textbf{(a) Decoder FD ERF:} perturb one latent site by $\varepsilon$ in all channels and plot $\|D(z+\varepsilon e)-D(z)\|_2$ per pixel (log). Visualization uses adaptive $\varepsilon$ ($\sim$0.5\% decoded RMS; clamp $[2{\times}10^{-3},10^{-2}]$); \emph{metrics} use fixed $\varepsilon{=}10^{-3}$. \textbf{Radii: $r_{50}\!\approx\!0.044\pm0.024$, $r_{90}\!\approx\!0.291\pm0.099$}. The bright core plus global low-amplitude ‘blanket’ highlights decoder non-locality, seems to echo high contrast structures in source image. \textbf{(b)~Encoder FD ERF:} inject a 1px impulse ($\delta{=}0.05$) at pixel $(i,j)$, compute $\Delta z{=}E(x+\delta e_{i,j}){-}E(x)$, and plot $\|\Delta z_{\cdot,u,v}\|_2$ per latent site (log). \textbf{Radii: $r_{50}\!\approx\!0.091\pm0.082$, $r_{90}\!\approx\!0.356\pm0.099$}. \textbf{(c) Gradient ERF:} for $y{=}D(z)$ and $s{=}\sum_{\text{5}\times\text{5}} y^2$, backpropagate to $\partial s/\partial z$ and show channelwise $\ell_2$ per latent site. Central peak and multiple secondary latent clusters relied on by the same pixel patch, exposing repeated structure. \textbf{(d) Energy curves:} Shaded region shows 95\% bootstrap confidence; both Encoder and Decoder shows a sharp core with long, low-amplitude tails, showing large ERFs; evidence that heuristic latent masking is inconsistent with the VAE and motivates DecFormer.}
    \label{fig:erf}
\end{figure}

\paragraph{Boundary leakage} Flux's VAE has a large receptive field, with global effects due to attention presence in the middle layer. Excluding the attention during mid-layers, we use recurrence formula defined by \cite{araujo2019computing} to find the encoder receptive field to be 217 pixels for each latent position and decoder to be 35.5 latents per pixel (Appendix \ref{sec:RF analysis}). For the decoder, we further calculate each latent's "influence field" to be 536 pixels in the reconstructed image. Measuring the effective receptive field through perturbations and gradient probes, the cumulative-energy radii $r_p$ demonstrates Flux's autoencoder to be sharply peaked and heavy tailed (Figure \ref{fig:erf}).  On a $256\times 256$ image, the encoder's $r_{90}\approx0.356=129$ pixels, and the decoder's $r_{90}\approx0.291=105$ pixels. In theoretical and experimental examinations we find that information not localized to the $8\times8$ grid. Thus, masking latents heuristically leads to both missing and leaked information.

\paragraph{Invalid interpolation} When heuristically compositing with soft masks, non-binary regions linearly interpolate two latents. However, we find that two latents cannot be mixed linearly to produce still valid latents. Empirically, more than half of voxels require mixing coefficients $\alpha$ outside $[0,1]$ to reproduce the ground-truth encoding. 

\paragraph{Mask downsampling} Resizing the pixel mask to latent resolution discards fine structure, as it downsamples the mask by $\frac{1}{8}$. In effect, heuristic masking cannot apply a finer mask than $\frac{1}{8}$ the resolution, which is inadequate for high-resolution inpainting.

Independent analyses of autoencoder interpolations~\citep{oring2021aeai} show that convex latent blends leave the data manifold, underscoring the need for an explicit compositor. Our contribution targets this missing piece: a lightweight latent operator that enforces pixel-equivalent blending, offloading mask geometry from giant finetunes while leaving semantic inpainting to those backbones.

\subsection{Related Work - LDM Inpainting and Editing}
Guided diffusion editing usually enlarges or retrains the denoiser. ControlNet adapters~\citep{zhang2023controlnet}, Paint-by-Example~\citep{yang2023paint}, DiffEditor~\citep{mou2024diffeditor}, BrushNet~\citep{ju2024brushnet}, and PowerPaint~\citep{zhuang2024powerpaint} each introduce hundreds of millions of new weights and require multi-GPU training just to teach the backbone mask awareness. In practice, foundation deployments still ship separate inpainting checkpoints—Stable Diffusion maintains SD-inpaint and SDXL-inpaint variants~\citep{sd15_inpaint_card,sdxl_inpaint_card}, while FLUX Fill is a second 12B rectified-flow model~\citep{blackforestlabs2024fluxfill}. The added maintenance cost, and often worse generation ability has spurred community forks that graft extra control nets on top of the main models in search of cleaner seams~\citep{alimama_flux_cn_github,alimama_flux_cn_hf,replicate_flux_cn,diffusers_flux_issue}. LatentPaint~\citep{latentpaint_wacv24} trims the training burden but still hands seam quality to the denoiser via broadcast latent masks.

\paragraph{Diffusion Inversion} Trajectory-level methods edit sampling paths instead: SDEdit~\citep{sdedit2022}, DiffEdit~\citep{couairon2023diffedit}, Blended Diffusion~\citep{avrahami2022blended}, and Differential Diffusion~\citep{levin2023differential} modulate the noise schedule or splice reference latents during denoising. These strategies improve controllability yet still stitch latents with a single-channel mask at each step, so VAE boundary coupling produces halos.

\section{Methodology}
\subsection{Formulation}

\paragraph{PELC objective}
We define PE for a unary latent operator via \eqref{eq:def_DE}, and specify latent composition as \eqref{eq:latent_composition}. We define the PELC objective to train latent compositor $C_F$:
Let $\hat{z} = C_F^\theta(z_1, z_2, M)$ be the predicted latent and $z_T = E(F(x_1, x_2, M))$ be the target latent. Let $\hat{x} = D(\hat{z})$ and $x_T = D(z_T)$ be their decodings.
\begin{align}
    \mathcal{L}_\text{PELC} &= \lambda_{E}\mathcal{L}_{\text{E}} + \mathcal{L}_{\text{D}}
    \label{eq:loss_formulation}
    \\[4pt]
    \mathcal{L}_{\text{E}} &= \mathbb{E} [ \| \hat{z} - z_T \|_2^2 ]
    \\[4pt]
    \mathcal{L}_{\text{D}} &= \mathbb{E} [ \mathcal{L}_{\text{LPIPS}}(\hat{x}, x_T) + \lambda_{H}\mathcal{L}_{\text{Halo-L1}}(\hat{x}, x_T) ]
\end{align}
Note that pixel supervision from the Encoded-Decoded target is important such that VAE-related errors cancel rather than contaminate the objective.  
\paragraph{Formulation}
Heurisic latent blending by broadcasting a downsampled pixel mask produces systematic errors: artifacts cluster at mask boundaries and leakage arises from globally entangled VAE channels (Figure~\ref{fig:sdf_combined}).  
Despite this, interpolation remains a strong inductive prior especially away from mask edges; we therefore instantiate PELC for inpainting as channel-adaptive---a hugely stronger formulation than broadcast masking---blending along the $z_A\!\leftrightarrow\!z_B$ line, augmented with an explicit residual correction.  
We predict
\[
\hat z = (1-\alpha)z_A + \alpha z_B + s, \hspace{0.5em}
\alpha \in [0,1]^{C\times H \times W}, s\in\mathbb{R}^{C\times H \times W}.
\]
Constraining $\alpha$ to $[0,1]$ yields a stable blend axis; $s$ absorbs orthogonal leakage and curvature. This separation is critical: if $\alpha$ is unconstrained it collapses both roles, destabilizing training. Appendix~\ref{app:projection} gives a closed-form decomposition into $(\alpha^*,s^*)$, and ablations are shown in Table~\ref{tab:ablations}.

\paragraph{Architecture}
Each block receives a \emph{rich feature stack}: latents $(z_A,z_B)$, the running $(\alpha,s)$, and error cues $\|z^{(t)}-z_A\|,\|z^{(t)}-z_B\|$. Re-patching/unpatching permits injection of these per-voxel errors at each stage. Multi-scale blocks ($[4,2,1,1]$ patch sizes) let coarse stages cheaply gather context, while patch $=1$ refines pixel-level boundaries. Local convolutions after unpatching also suppress 1--2px halos. This design is both FLOP-efficient and effective at feeding both global and local signals. A lightweight, two-headed CNN, run only once-per-mask, maps the $M$ to a content-agnostic prior $\alpha_0$, seeding the blend and block 0 inputs, and provides embeddings for cross-attention.
\paragraph{Conditioning}
Blended latent errors concentrate near mask edges due to entangled channels; we therefore compute a pixel and latent-sized halo: a softly decaying band ($\approx$8 px radius in pixel space, stride-scaled, motivated by results from Figure~\ref{fig:sdf_combined} and Appendix~\ref{alphashiftinternals}) around mask boundaries and soft regions. The halo serves two roles. First, it directly conditions the model (via FiLM) so that both $\alpha$ and $s$ are aware of boundary context. Second, it provides a loss weighting that emphasizes precisely those regions where naïve interpolation fails.
Cross-attention to mask tokens is confined to the patch $=1$ blocks. Global blocks already access coarse mask embeddings through FiLM; but fine-scale editing at pixel resolution requires precise spatial alignment, which attention provides. Restricting cross-attention to the final stage keeps compute low while preserving boundary fidelity. FiLM also recieves input from a latent sized mask image, which gives strong signal to the alpha head.

\paragraph{Alpha-Shift separation}
To disentangle the roles of $\alpha$ and $s$, we experimented with several explicit regularizers:  
(i) a scale-aware $L_1$ penalty on $s$ controlled by an EMA target magnitude,  
(ii) a cosine-hinge that penalizes $|\cos(s,d)|$ when aligned with $d=z_A-z_B$, and  
(iii) direct supervision against $(\alpha^*,s^*)$.  
However, we found that such constraints often over-regularized the model and hindered convergence.  
Instead, we adopt a staged training schedule: the shift head remains gated off until $\alpha$ has converged, at which point $s$ is warmed up gradually (see Appx.~\ref{sec:staging}). 
Associated losses, including the halo-weighted $L_1$, are likewise ramped in during this phase, ensuring that $\alpha$ receives a clean learning signal early on—particularly important since $\alpha$ alone cannot correct decoder leakage at mask boundaries.

\paragraph{Efficiency and FLOP Budget}
The full module has 7.7M parameters, almost three orders of magnitude fewer than the 12B-parameter diffusion model it augments. At 1024×1024 resolution, DecFormer has 7.0M parameters and costs 80.4 GFLOPs per diffusion step, while the mask pre-processor has 0.7M parameters and costs 28.5 GFLOPs (run once per mask or diffusion generation). For a standard 28-step generation, this adds only $\approx$2.3 TFLOPs in total, corresponding to a 3.4\% overhead on top of FLUX’s 66-TFLOP 1024×1024 generation.
\paragraph{Training}
We train DecFormer on NVIDIA H100 GPUs. Each run uses a batch size of 8 and proceeds for $ 8\times 10^4 $ steps (approx 128 epochs, $ \sim\!10^6 $ updates in total).  
Inputs are sampled at multi-resolution from $256{\times}256$ to $384{\times}384$ pixels with aspect ratios in $[0.5,2.0]$.  
Mask augmentation includes graduated edge detection (0--15\%) and feathering ramps over 1000 steps.  
Optimization employs AdamW with cosine SGDR: warm restarts at 4k, 12k, and 14k steps with $\eta_{\max}=10^{-3}$, $\eta_{\min}=2{\times}10^{-4}$, followed by cosine annealing from 30k to 60k steps down to $10^{-4}$.
\paragraph{Data}
We train DecFormer on a heterogeneous mix of images and masks.  
Our image set combines 30k natural photographs from Flickr30k, 10k artworks from WikiArt, and an additional 100k high-resolution images collected from internal web sources.  
Conditioning masks are drawn from P3M\cite{li2021privacypreservingportraitmatting}, GFM \cite{li2021bridgingcompositerealendtoend}, and procedurally generated random shapes.

\begin{figure}
    \centering
    \includegraphics[width=\linewidth]{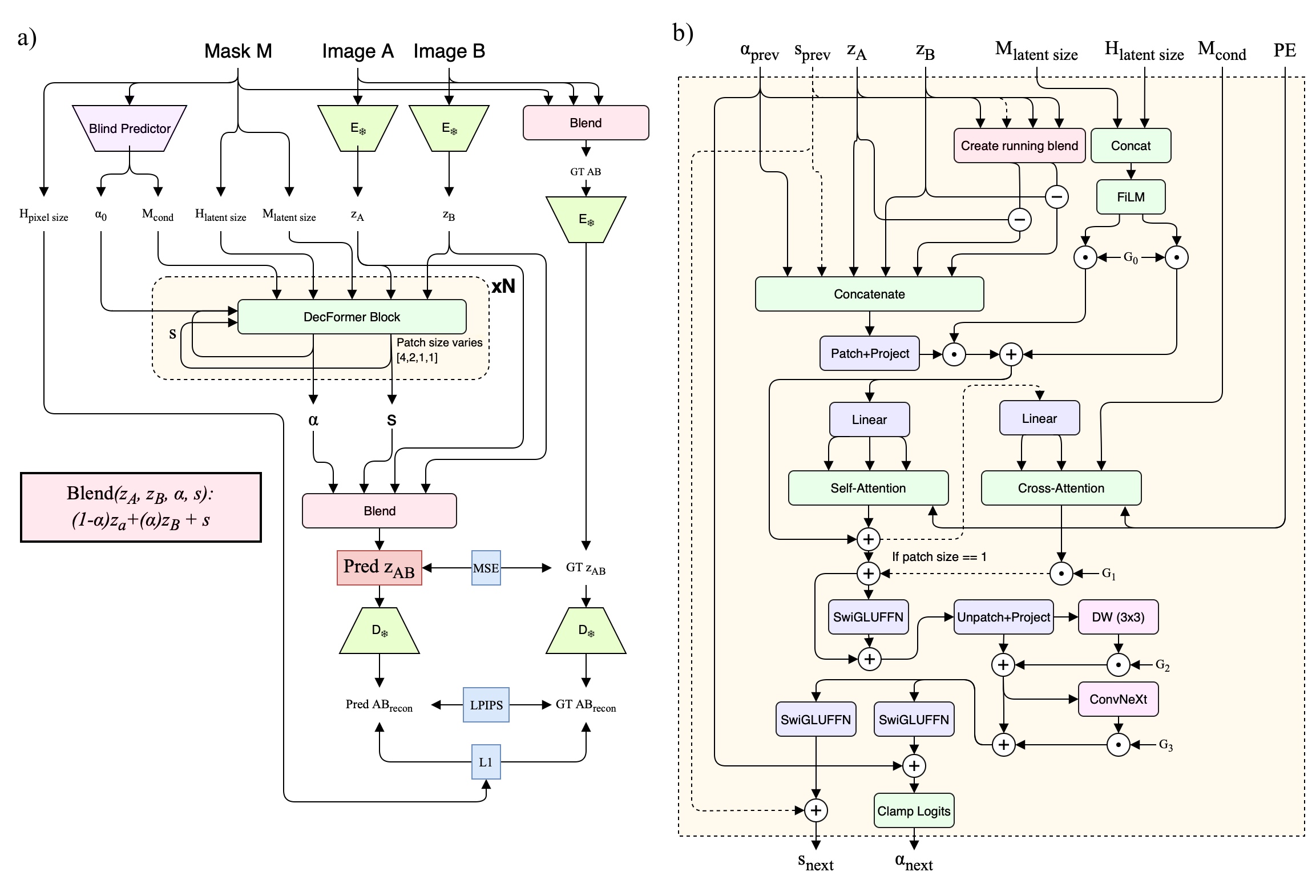}
    \caption{Overview of our training pipeline and DecFormer architecture. The left panel illustrates the overall flow: two input images and a pixel mask are encoded by a frozen VAE, the mask is processed by a lightweight CNN prior (architecture detailed in Appendix \ref{blind-pred-arch}), and DecFormer predicts channel-adaptive blend weights $\alpha$ and residual corrections $s$ at latent resolution. The right panel zooms into a single DecFormer block, showing the feature stack, patching/unpatching, FiLM conditioning, attention and cross attention. For an expanded diagram on the DecFormer architecture see Appendix~\ref{deltaformer-extended-arch}.}
    \label{fig:decformer-architecture}
\end{figure}

\subsection{Using DecFormer for Diffusion Inpainting}
\label{sec:scheduler}

DecFormer is trained on unnoised latents ($z_0$), so we reformulate the scheduler step to predict and blend at the fully denoised $z_0$ then re-noise to $z_{t-1}$. We \textbf{(A)} decode to $z_0$, \textbf{(B)} blend there, then \textbf{(C)} re-target the velocity to land on the composed $z_0^\star$ and step once:
\[
\begin{aligned}
&\textbf{A:}\quad z_0^\theta \;=\; z_t - t\, v_\theta(z_t, t) \\[4pt]
&\textbf{B:}\quad 
z_0^\star \;=\;
(1-\alpha)\odot z_0^\theta + \alpha\odot z_0^{\text{ref}} + s\\[8pt]
&\textbf{C:}\quad v^\star \;=\; \dfrac{z_t - z_0^\star}{t}, \qquad
z_{t'} \;=\; z_t + (t' - t)\, v^\star.
\end{aligned}
\]
For \textbf{(B)}, we predict $(\alpha,s)=\mathcal{D}_\phi(z_0^\theta, z_0^{\text{ref}}, M)$ from trained model $\mathcal{D}_\phi$. DecFormer acts as a direct replacement for heuristic blending during sampling.

\section{Experiments}
\label{sec:experiments}

\subsection{Network Design Ablations}

\begin{table}[h]
\centering
\caption{Ablation results. Removing halo-focused losses degrades boundary quality ($\uparrow$Halo L1) even though global metrics are competitive. Removing the residual shift head (unconstrained $\alpha$-only) substantially harms all metrics, confirming the need for both $\alpha$ and residual $s$. Baseline balances both.}
\label{tab:ablations}
\resizebox{\columnwidth}{!}{%
\begin{tabular}{lccc}
\toprule
Experiment & Halo L1 ($\downarrow$) & LPIPS ($\downarrow$) & MSE ($\downarrow$) \\
\midrule
No Halo L1 Loss & $0.0973 \pm 0.0002$ & \textbf{$0.0299 \pm 0.0003$} & \textbf{$0.0297 \pm 0.0003$} \\
Baseline & \textbf{$0.0829 \pm 0.0018$} & $0.0303 \pm 0.0015$ & $0.0303 \pm 0.0003$ \\
Unconstrained Alpha, No Shift & $0.1079 \pm 0.0012$ & $0.0514 \pm 0.0012$ & $0.0331 \pm 0.0003$ \\
\bottomrule
\end{tabular}
}
\end{table}

We tested targeted ablations to isolate which components materially contribute to pixel-consistent compositing.  
Table~\ref{tab:ablations} reports dataset-level means $\pm$95\% CIs over three seeds up to 80,000 steps.

\paragraph{Ablation strategy}
We prioritize our presented ablations to test the principles that make equivalence attainable. We fully train ablated models with 80k steps, using three seeds. We report dataset-level means with 95\% CIs for halo-weighted L1, LPIPS \cite{zhang2018unreasonableeffectivenessdeepfeatures}, and MSE.

\subsection{DecFormer Achieves Pixel-Equivalent Compositing}
\paragraph{Metrics}We evaluate DecFormer by comparing its reconstruction fidelity against heuristic masking as the baseline. We evaluate on images from the Coco 2017 validation set and masks from the Compositions-1k dataset. We augment the masks in 3 ways to evaluate composition quality for specific mask types: Gaussian blurred soft masks with sigma=21, binarized masks, and thin masks by extracting the original mask edges. We use set seeds and dataset hashing to ensure the same experimental set is used each time.

\paragraph{Results}Our results show a decisive improvement across all metrics and mask types at all resolutions (Table~\ref{tab:decformer-interp-metrics}, extra resolutions and comparisons to different naive downscaling methods are shown in Appendix~\ref{extended-metrics}). The results further show that heuristic blending introduces measurable artifacts that DecFormer minimizes.

\begin{figure*}[htbp]
    \centering
    \includegraphics[width=0.8\textwidth]{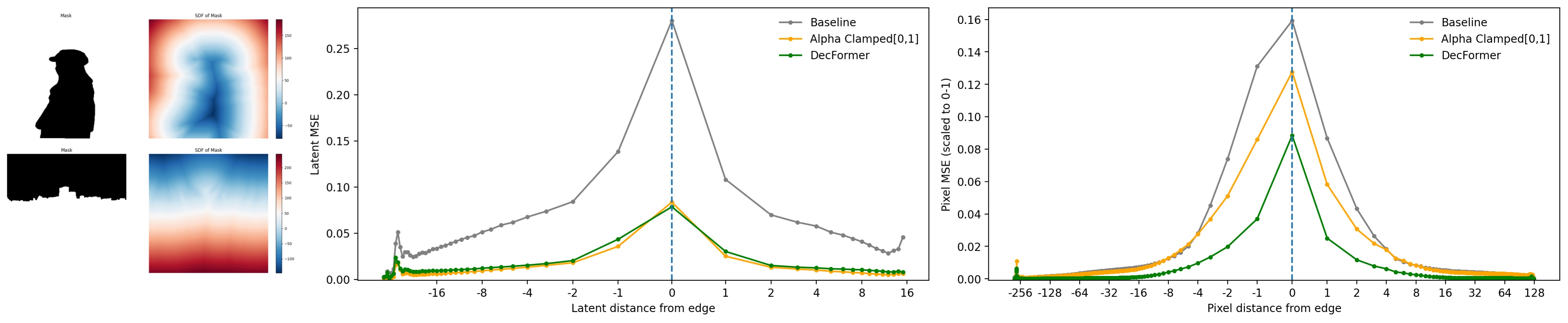}
    \caption{Signed-distance analysis of mask edges. Left: example masks and their signed distance fields (SDF). Middle: Mean per-latent MSE relative to latent downsized mask SDF. Right: Mean per-pixel MSE relative to mask SDF. Heuristic (grey) applies the downsampled binary mask in latent space; Alpha‑clamped (orange) uses the clamped least-squares solved $\alpha^*$ Appendix~\ref{app:projection}; DecFormer (green) is our model. DecFormer achieves the lowest error around the boundary with a sharper fall-off compared to baselines.}
    \label{fig:sdf_combined}
\end{figure*}

\paragraph{Signed Distance Field Analysis} To spatially locate where these improvements originate, we analyze the reconstruction error in relation to the signed distance field (SDF) from the mask boundary, shown in Figure~\ref{fig:sdf_combined}. For a given mask, each pixel's signed distance is defined as the orthogonal distance from the mask boundary, with negative values indicating the pixel being inside the mask and positive values for outside the mask. 

We evaluate per-pixel and per-latent MSE for 10k images and masks, and for each distance value, we plot the mean MSE. The heuristic baseline (grey) exhibits a prominent error spike at the mask boundary (distance = 0) that decays slowly into both the inside and outside of mask regions.  In contrast, DecFormer (green) achieves a noticeably lower peak error and sharper error fall-off. 

\paragraph{Visual Comparison}These quantitative improvements translate directly to superior visual quality. As illustrated in qualitative comparisons in Figure \ref{fig:hero} (and more extensively in Appendix \ref{hero-extended}), DecFormer produces seamless composites free of the halos, color shifts, and jagged boundary artifacts endemic to heuristic masking.
We create a detailed visualization of DecFormer’s internal predictions for the channel-wise blend weights ($\alpha$) and the residual correction ($s$), included in Appendix \ref{alphashiftinternals}. Decformer utilizes the residual correction extensively at mask edges and soft mask areas, demonstrating that our training strategy leads to the model relying on convex blending where possible, and using the residual correction at challenging areas.

\begin{table}[t]
\centering
\caption{Comparison of DecFormer and the heuristic baseline at 1024px resolution (mean ± 95\% CI, n=50).}
\label{tab:decformer-interp-metrics}
\resizebox{\columnwidth}{!}{%
\begin{tabular}{llcccc}
\toprule
\textbf{Mask Type} & \textbf{Method} & \textbf{SSIM} $\uparrow$ & \textbf{PSNR (dB)} $\uparrow$ & \textbf{LPIPS} $\downarrow$ & \textbf{Halo L1} $\downarrow$ \\
\midrule
\multirow{2}{*}{Soft ($\sigma$=21)} & DecFormer & \textbf{0.985}$_{\pm.003}$ & \textbf{41.3}$_{\pm.8}$ & \textbf{0.027}$_{\pm.005}$ & \textbf{0.018}$_{\pm.001}$ \\
 & Heuristic & 0.941$_{\pm.010}$ & 32.9$_{\pm1.1}$ & 0.088$_{\pm.016}$ & 0.050$_{\pm.005}$ \\
\cmidrule(l){2-6}
\multirow{2}{*}{Binary} & DecFormer & \textbf{0.964}$_{\pm.017}$ & \textbf{35.7}$_{\pm1.5}$ & \textbf{0.045}$_{\pm.018}$ & \textbf{0.060}$_{\pm.006}$ \\
 & Heuristic & 0.913$_{\pm.025}$ & 28.4$_{\pm1.3}$ & 0.110$_{\pm.029}$ & 0.141$_{\pm.008}$ \\
\cmidrule(l){2-6}
\multirow{2}{*}{Original} & DecFormer & \textbf{0.968}$_{\pm.016}$ & \textbf{38.6}$_{\pm1.5}$ & \textbf{0.049}$_{\pm.018}$ & \textbf{0.037}$_{\pm.005}$ \\
 & Heuristic & 0.918$_{\pm.024}$ & 31.1$_{\pm1.4}$ & 0.104$_{\pm.028}$ & 0.080$_{\pm.007}$ \\
\cmidrule(l){2-6}
\multirow{2}{*}{Thin} & DecFormer & \textbf{0.967}$_{\pm.014}$ & \textbf{34.7}$_{\pm1.5}$ & \textbf{0.045}$_{\pm.017}$ & \textbf{0.073}$_{\pm.005}$ \\
 & Heuristic & 0.920$_{\pm.030}$ & 27.3$_{\pm1.2}$ & 0.111$_{\pm.031}$ & 0.174$_{\pm.009}$ \\
\bottomrule
\end{tabular}
}
\end{table}

\subsection{DecFormer is a strong Diffusion Inpainting Prior}
\paragraph{Setup}We now evaluate DecFormer as an inpainting prior for a diffusion backbone, both frozen and tuned with a small inpainting LoRA \cite{hu2021loralowrankadaptationlarge}. The LoRA uses a dual-sigma noise schedule (Appendix~\ref{lora}) and is trained independently to demonstrate complementary benefits with DecFormer. We hypothesize that repairing the masking operation during the diffusion denoising process yields end-to-end improvements for a masked generation task. We integrate DecFormer into Flux.1-Dev using the $z_0$ re-targeted scheduler from \S\ref{sec:scheduler}. At each sampling step (for 30 steps), we update the predicted velocity $v^\star$ using mask $m$ and reference $z_0^{\text{ref}}$. For training the LoRA, we use COCO-2017 train images with instance segmentations as masks. For evaluation, we report metrics on the validation split, filtering examples with masked area $>15\%$ of the image; prompts and guidance are held fixed across methods.

\paragraph{Results}
We compare five variants:
\begin{enumerate}
\item \textbf{Heuristic}: downsampled, broadcast latent mask blended at every step (Equation\eqref{eq:heuristic_masking_latent}).
\item \textbf{DecFormer}: our compositor inserted at every step via $z_0$ re-targeting; backbone frozen.
\item \textbf{LoRA only}: LoRA on \texttt{Flux.1-Dev} trained for inpainting using the same masks; no compositor.
\item \textbf{DecFormer + LoRA}: DecFormer plus the inpainting LoRA.
\item \textbf{Flux.1-Fill}: a fully finetuned, mask-aware inpainting model used as a strong reference.
\end{enumerate}

\begin{figure*}[htbp]
    \centering
    \includegraphics[width=0.75\textwidth]{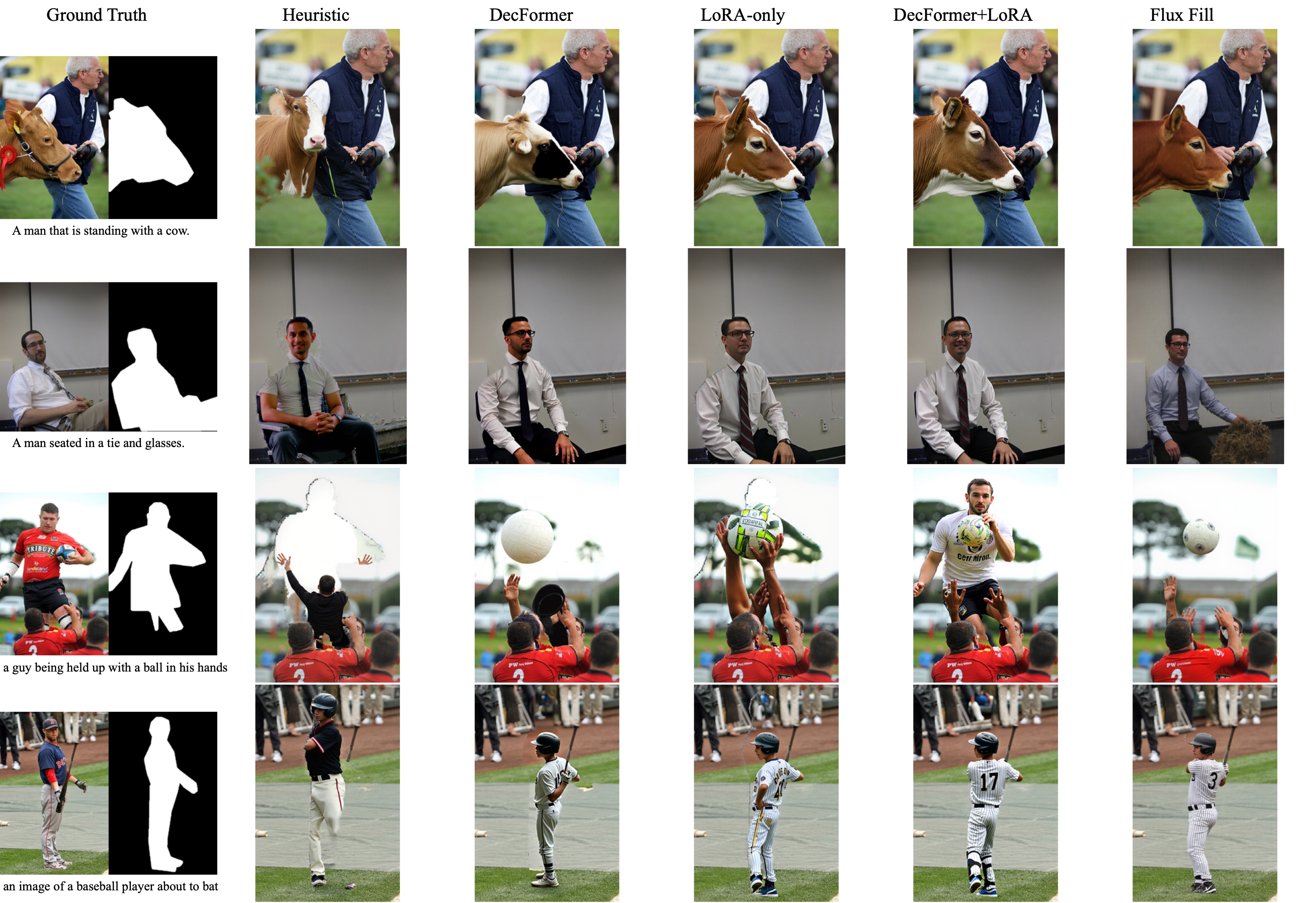}
    \caption{Inpainting quality comparisons for 4 images between heuristic, DecFormer, light-training inpainting LoRA, LoRA with DecFormer, and Flux.1 Fill, a fully finetuned image editing model.}
    \label{fig:inpainting_visual}
\end{figure*}

We report SSIM, PSNR, LPIPS, and FID against ground-truth mask composites. DecFormer, without any backbone finetuning, improves inpainting over the latent-heuristic baseline across all metrics (see Table ~\ref{table:inpainting_results} and qualitative examples in Fig.~\ref{fig:inpainting_visual}).

Adding a small LoRA (\emph{DecFormer+LoRA}) further closes the gap to a  dedicated inpainting model. \emph{DecFormer+LoRA} achieves perceptual quality comparable to the fully  finetuned Flux.1-Fill, though pixel-level reconstruction remains weaker (Table~\ref{table:inpainting_results}).

\paragraph{Visual Comparison}Figure~\ref{fig:inpainting_visual} illustrates common failure modes with the heuristic. DecFormer reduces boundary artifacts, and the edits are more aware of the masked area. LoRA primarily helps with semantic plausibility inside the masked region, whereas DecFormer governs how content is stitched, yielding complementary gains.

We show that errors introduced by heuristic masking indeed impact the generation quality for inpainting, and by correcting these errors with DecFormer, we gain inpainting quality improvements for training-free and light training contexts.

\begin{table}[h!]
\centering
\caption{Masked Editing results. Evaluated on all samples with masked area $>$ 0.15 in COCO-2017 validation set. }
\resizebox{\columnwidth}{!}{%
\begin{tabular}{l c c c c}
\hline
Method & SSIM (mean$\pm$std) & PSNR (mean$\pm$std) & LPIPS (mean$\pm$std) & FID \\
\hline
Baseline (Heuristic) & 0.643 $\pm$ 0.145 & 13.578 $\pm$ 2.915 & 0.354 $\pm$ 0.152 & 23.514 \\
DecFormer                & \textbf{0.682 $\pm$ 0.139} & 13.943 $\pm$ 2.870 & 0.314 $\pm$ 0.143 & 20.556 \\
LoRA with no DecFormer   & 0.653 $\pm$ 0.142 & 14.160 $\pm$ 2.620 & 0.331 $\pm$ 0.143 & 21.519 \\
Flux Fill            & 0.681 $\pm$ 0.141 & \textbf{16.750 $\pm$ 3.199} & 0.313 $\pm$ 0.125 & 19.343 \\
DecFormer with LoRA      & 0.680 $\pm$ 0.139 & 14.231 $\pm$ 2.742 & \textbf{0.303 $\pm$ 0.138} & \textbf{19.280} \\
\hline
\end{tabular}
}

\label{table:inpainting_results}
\end{table}

\subsection{Proof of Generality beyond Compositing}
\label{ssec:proof_of_concept}

\paragraph{Objective}
To validate the generality of the PELC objective beyond multi-image tasks, we test its ability to learn a complex parametric color transformation. This operator, defined as $F(x; \gamma, c, b) = (x^{1/\gamma} - 0.5) \cdot c + 0.5 + b$, combines gamma, contrast, and brightness adjustments. We compare LPIPS, PSNR, and SSIM metrics between our model and applying $F$ directly onto latents as a baseline.
\paragraph{Setup \& Results}
We train a lightweight, FiLM-conditioned transformer to predict a latent residual, conditioned on the transformation parameters $(\log \gamma, \log c, b)$. The model is optimized via the PELC objective to be pixel-equivalent, using a combined objective of latent MSE and pixel-space LPIPS. As demonstrated quantitatively in Table~\ref{fig:gamma_validation_vertical} and qualitatively in Appendix~\ref{non-compositing-img}, our PELC-trained model successfully learns this complex mapping, reproducing the target transformation with high fidelity, whereas naively operating on the latent causes the image to degrade catastrophically. This success validates that PELC is a general framework for learning pixel-equivalent latent operators, including single-image parametric adjustments. We report results n=1024 on the COCO-2017 validation set.

\begin{table}[t]
\centering
\caption{
    Quantitative Metrics (Mean $\pm$ 95\% CI) on 1024 random samples images from the COCO-2017 validation dataset. The baseline applies the parametric color-transformation formula directly in latent space, following the same simplifying assumption used in heuristic latent masking.
}
\label{fig:gamma_validation_vertical}

    \centering
    \sisetup{separate-uncertainty=true}
    \begin{tabular}{@{}l S[table-format=1.4(4)] S[table-format=2.4(4)]@{}}
    \toprule
    Metric & {Baseline (heuristic)} & {PELC-trained model (Ours)} \\
    \midrule
    LPIPS $\downarrow$ & 0.4996 \pm 0.0076 & \bfseries 0.0875 \pm 0.0023 \\
    PSNR $\uparrow$  & 18.1630 \pm 0.1968 & \bfseries 27.2835 \pm 0.2301 \\
    SSIM $\uparrow$  & 0.4359 \pm 0.0112 & \bfseries 0.8466 \pm 0.0059 \\
    \bottomrule
    \end{tabular}
    \label{fig:gamma_table_vertical}

\end{table}

\section{Conclusion}
Linear latent mixing treats latents as pseudo-pixels, but modern VAEs are nonlinear and globally entangled, breaking pixel-equivalence and producing boundary and color artifacts that sampling amplifies. Pixel-Equivalent Latent Compositing (PELC) provides a principled alternative: we learn pixel-consistent latent operators using a frozen VAE for pixel-space supervision, avoiding the drift and computational cost of per-step encode–decode cycles.

DecFormer instantiates this idea with a lightweight transformer that predicts per-channel blend weights and a nonlinear residual. It preserves thin structures, soft opacity gradients, and high-resolution boundaries that broadcast masking cannot represent after VAE downsampling. DecFormer reduces boundary errors by up to 53\% and halves perceptual error (LPIPS) compared to heuristic masking (\S\ref{sec:experiments}) with negligible overhead (7.7M parameters, $<$0.1\% of backbone size, 3.5\% compute overhead).

Because DecFormer corrects \emph{how} latents are fused rather than \emph{what} is generated, its benefits are orthogonal to semantic reasoning. It composes cleanly with mask-aware LoRAs and approaches the quality of fully finetuned inpainting backbones while remaining far smaller and cheaper than ControlNet-style hypernetworks. Beyond compositing, controlled experiments on nonlinear color transforms (\S\ref{ssec:proof_of_concept}) show that pixel-equivalence extends naturally to non-compositing latent edits as well.

\paragraph{Limitations and future work}
PELC addresses fusion, not semantic reconstruction; large context-dependent edits still require mask-aware denoisers. Broader validation on additional VAEs is required. Extending PELC to additional latent operators, including spatial warps and temporally coherent video edits, is a natural next step as well as supporting work investigating how training inpainting models or ControlNets with PELC in the loop could reduce task difficulty speed convergence or network size requirements.

Pixel-equivalent latent compositing exposes a foundational flaw in current inpainting practice and replaces it with a simple, general, and geometrically consistent principle. It yields sharper seams, correct opacity handling, and high-resolution boundary fidelity with minimal cost, opening a path toward a library of decode-consistent latent operators for modern diffusion pipelines.

\FloatBarrier

\section*{Author Contributions}
R.B. initiated the research, formulated the core inconsistency of latent masking, developed the decoder-equivalence principle, DecFormer architecture, and training objectives. D.Z. built the training infrastructure and curated all datasets. Both jointly refined the PELC objective; D.Z. formulated the encoder–decoder pixel-equivalence extension from decoder-equivalence. R.B. performed the ERF analyses, developed the $z_0$-retargeting formulation and the $\gamma$-correction generality, and led DecFormer training, ablations, and evaluation. D.Z. led all VAE latent-space analyses, designed the dual-sigmas inpainting schedule, and performed the diffusion and inpainting experiments, including FLUX/FLUX-Fill finetuning and evaluation. Both authors contributed equally to writing, visualization, and manuscript preparation. R.B. arranged computing resources from Modal that enabled this work.

\section*{Acknowledgements}
This work was conducted by the Bradbury Group, an independent non-profit AI research lab. We gratefully acknowledge Modal for providing the compute support that enabled our large-scale parallel training. Beyond the listed authors, we thank the broader Bradbury Group research community for internal reviews, discussion of early prototypes, and support with maintaining our shared infrastructure.

\bibliography{bibliography}

@inproceedings{rombach2022high,
  title     = {High-Resolution Image Synthesis with Latent Diffusion Models},
  author    = {Rombach, Robin and Blattmann, Andreas and Lorenz, Dominik and Esser, Patrick and Ommer, Bj{\"o}rn},
  booktitle = {Proceedings of the IEEE/CVF Conference on Computer Vision and Pattern Recognition (CVPR)},
  pages     = {10684--10695},
  year      = {2022},
  url       = {https://arxiv.org/abs/2112.10752}
}

@misc{blackforestlabs2024fluxfill,
  author       = {{Black Forest Labs}},
  title        = {FLUX.1 Fill: State-of-the-art inpainting and outpainting models},
  year         = {2024},
  month        = {November},
  howpublished = {\url{https://huggingface.co/black-forest-labs/FLUX.1-Fill-dev}},
  note         = {12B parameter inpainting model}
}

@misc{sd15_inpaint_card,
  title        = {Stable Diffusion Inpainting Model Card (v1.5)},
  author       = {{RunwayML}},
  howpublished = {\url{https://huggingface.co/stable-diffusion-v1-5/stable-diffusion-inpainting}},
  year         = {2022},
  note         = {UNet has 5 additional input channels (4 masked-image latents + 1 mask)}
}

@misc{sdxl_inpaint_card,
  title        = {Stable Diffusion XL 1.0 Inpainting 0.1},
  howpublished = {\url{https://huggingface.co/diffusers/stable-diffusion-xl-1.0-inpainting-0.1}},
  year         = {2023},
  note         = {Adopts the same 5 additional input channels for inpainting}
}

@article{couairon2023diffedit,
  title   = {DiffEdit: Diffusion-based semantic image editing with mask guidance},
  author  = {Couairon, Guillaume and Verbeek, Jakob and Schwenk, Holger and Cord, Matthieu},
  journal = {International Conference on Learning Representations (ICLR)},
  year    = {2023},
  url     = {https://arxiv.org/abs/2210.11427}
}

@article{sdedit2022,
  title   = {SDEdit: Guided Image Synthesis and Editing with Stochastic Differential Equations},
  author  = {Meng, Chenlin and He, Yutong and Song, Yang and Song, Jiaming and Wu, Jiajun and Zhu, Jun-Yan and Ermon, Stefano},
  journal = {International Conference on Learning Representations (ICLR)},
  year    = {2022},
  url     = {https://arxiv.org/abs/2108.01073}
}

@inproceedings{avrahami2022blended,
  title     = {Blended Diffusion for Text-Driven Editing of Natural Images},
  author    = {Avrahami, Omri and Lischinski, Dani and Fried, Ohad},
  booktitle = {Proceedings of the IEEE/CVF Conference on Computer Vision and Pattern Recognition (CVPR)},
  pages     = {18208--18218},
  year      = {2022},
  url       = {https://arxiv.org/abs/2111.14818}
}

@inproceedings{yang2023paint,
  title     = {Paint by Example: Exemplar-Based Image Editing with Diffusion Models},
  author    = {Yang, Binxin and Gu, Shuyang and Zhang, Bo and Zhang, Ting and Chen, Xuejin and Sun, Xiaoyan and Chen, Dong and Wen, Fang},
  booktitle = {Proceedings of the IEEE/CVF Conference on Computer Vision and Pattern Recognition (CVPR)},
  pages     = {18381--18391},
  year      = {2023},
  url       = {https://arxiv.org/abs/2211.13227}
}

@InProceedings{latentpaint_wacv24,
    author    = {Corneanu, Ciprian and Gadde, Raghudeep and Martinez, Aleix M.},
    title     = {LatentPaint: Image Inpainting in Latent Space With Diffusion Models},
    booktitle = {Proceedings of the IEEE/CVF Winter Conference on Applications of Computer Vision (WACV)},
    month     = {January},
    year      = {2024},
    pages     = {4334-4343}
}

@inproceedings{zhang2023controlnet,
  title     = {Adding Conditional Control to Text-to-Image Diffusion Models},
  author    = {Zhang, Lvmin and Rao, Anyi and Agrawala, Maneesh},
  booktitle = {Proceedings of the IEEE/CVF International Conference on Computer Vision (ICCV)},
  pages     = {3836--3847},
  year      = {2023},
  url       = {https://arxiv.org/abs/2302.05543}
}

@article{ju2024brushnet,
  title   = {BrushNet: A Plug-and-Play Image Inpainting Model with Decomposed Dual-Branch Diffusion},
  author  = {Ju, Xuan and Liu, Xian and Wang, Xintao and Bian, Yuxuan and Shan, Ying and Xu, Qiang},
  journal = {arXiv preprint arXiv:2403.06976},
  year    = {2024},
  url     = {https://arxiv.org/abs/2403.06976}
}

@article{zhuang2024powerpaint,
  title   = {A Task is Worth One Word: Learning with Task Prompts for High-Quality Versatile Image Inpainting},
  author  = {Zhuang, Junhao and Zeng, Yanhong and Liu, Wenran and Yuan, Chun and Chen, Kai},
  journal = {arXiv preprint arXiv:2312.03594},
  year    = {2024},
  url     = {https://arxiv.org/abs/2312.03594}
}

@inproceedings{oring2021aeai,
  title     = {Autoencoder Image Interpolation by Shaping the Latent Space},
  author    = {Oring, Alon and Yakhini, Zohar and Hel-Or, Yacov},
  booktitle = {Proceedings of the 38th International Conference on Machine Learning (ICML)},
  pages     = {8281--8290},
  year      = {2021},
  organization = {PMLR},
  url       = {https://proceedings.mlr.press/v139/oring21a.html}
}

@misc{alimama_flux_cn_github,
  title        = {FLUX-Controlnet-Inpainting},
  author       = {{Alimama Creative Team}},
  howpublished = {GitHub: \url{https://github.com/alimama-creative/FLUX-Controlnet-Inpainting}},
  year         = {2024}
}

@misc{alimama_flux_cn_hf,
  title        = {FLUX.1-dev ControlNet Inpainting (Alpha/Beta)},
  author       = {{Alimama Creative Team}},
  howpublished = {\url{https://huggingface.co/alimama-creative/FLUX.1-dev-Controlnet-Inpainting-Beta}},
  year         = {2024}
}

@misc{replicate_flux_cn,
  title        = {Flux-dev Inpainting ControlNet},
  howpublished = {Replicate: \url{https://replicate.com/zsxkib/flux-dev-inpainting-controlnet}},
  year         = {2024}
}

@misc{diffusers_flux_issue,
  title        = {Problem with FluxInpaintPipeline when doing a replace},
  howpublished = {Hugging Face Diffusers Issue \#9486: \url{https://github.com/huggingface/diffusers/issues/9486}},
  year         = {2024}
}

@misc{mou2024diffeditor,
      title={DiffEditor: Boosting Accuracy and Flexibility on Diffusion-based Image Editing}, 
      author={Chong Mou and Xintao Wang and Jiechong Song and Ying Shan and Jian Zhang},
      year={2024},
      eprint={2402.02583},
      archivePrefix={arXiv},
      primaryClass={cs.CV},
      url={https://arxiv.org/abs/2402.02583}, 
}

@article{levin2023differential,
  title   = {Differential Diffusion: Giving Each Pixel Its Strength},
  author  = {Levin, Eyal and Fried, Ohad},
  journal = {Computer Graphics Forum},
  volume  = {42},
  number  = {6},
  pages   = {e15040},
  year    = {2023},
  url     = {https://arxiv.org/abs/2306.00950}
}

@article{Peebles2022DiT,
  title={Scalable Diffusion Models with Transformers},
  author={William Peebles and Saining Xie},
  year={2022},
  journal={arXiv preprint arXiv:2212.09748},
}

@misc{bfl2025flux1kontextflowmatching,
      title={FLUX.1 Kontext: Flow Matching for In-Context Image Generation and Editing in Latent Space}, 
      author={Black Forest Labs and Stephen Batifol and Andreas Blattmann and Frederic Boesel and Saksham Consul and Cyril Diagne and Tim Dockhorn and Jack English and Zion English and Patrick Esser and Sumith Kulal and Kyle Lacey and Yam Levi and Cheng Li and Dominik Lorenz and Jonas Müller and Dustin Podell and Robin Rombach and Harry Saini and Axel Sauer and Luke Smith},
      year={2025},
      eprint={2506.15742},
      archivePrefix={arXiv},
      primaryClass={cs.GR},
      url={https://arxiv.org/abs/2506.15742}, 
}

@article{araujo2019computing,
  author = {Araujo, André and Norris, Wade and Sim, Jack},
  title = {Computing Receptive Fields of Convolutional Neural Networks},
  journal = {Distill},
  year = {2019},
  note = {https://distill.pub/2019/computing-receptive-fields},
  doi = {10.23915/distill.00021}
}

@misc{hu2021loralowrankadaptationlarge,
      title={LoRA: Low-Rank Adaptation of Large Language Models}, 
      author={Edward J. Hu and Yelong Shen and Phillip Wallis and Zeyuan Allen-Zhu and Yuanzhi Li and Shean Wang and Lu Wang and Weizhu Chen},
      year={2021},
      eprint={2106.09685},
      archivePrefix={arXiv},
      primaryClass={cs.CL},
      url={https://arxiv.org/abs/2106.09685}, 
}

@misc{li2021bridgingcompositerealendtoend,
      title={Bridging Composite and Real: Towards End-to-end Deep Image Matting}, 
      author={Jizhizi Li and Jing Zhang and Stephen J. Maybank and Dacheng Tao},
      year={2021},
      eprint={2010.16188},
      archivePrefix={arXiv},
      primaryClass={cs.CV},
      url={https://arxiv.org/abs/2010.16188}, 
}

@misc{li2021privacypreservingportraitmatting,
      title={Privacy-Preserving Portrait Matting}, 
      author={Jizhizi Li and Sihan Ma and Jing Zhang and Dacheng Tao},
      year={2021},
      eprint={2104.14222},
      archivePrefix={arXiv},
      primaryClass={cs.CV},
      url={https://arxiv.org/abs/2104.14222}, 
}

@misc{zhang2018unreasonableeffectivenessdeepfeatures,
      title={The Unreasonable Effectiveness of Deep Features as a Perceptual Metric}, 
      author={Richard Zhang and Phillip Isola and Alexei A. Efros and Eli Shechtman and Oliver Wang},
      year={2018},
      eprint={1801.03924},
      archivePrefix={arXiv},
      primaryClass={cs.CV},
      url={https://arxiv.org/abs/1801.03924}, 
}
\bibliographystyle{ieeenat_fullname}
\newpage

\FloatBarrier
\appendix
\section*{Appendix}
\section{Least-squares projection}
\label{app:projection}

Given target latent $z_T$ and source latents $z_A, z_B$, we seek the optimal decomposition:
\begin{align}
(\alpha^*, s^*) = \arg\min_{\alpha, s} &\;\; \big\|z_T - \big[(1-\alpha) z_A + \alpha z_B + s\big]\big\|_2^2 \\
\text{s.t.}\quad & \alpha \in [0,1]^{C \times H \times W}.
\end{align}

This yields the closed-form solution
\begin{align}
\alpha^*_{ijc} &= \Pi_{[0,1]}\!\left(\frac{(z_T^{ijc} - z_B^{ijc})(z_A^{ijc} - z_B^{ijc})}{\|z_A^{ijc} - z_B^{ijc}\|_2^2 + \epsilon}\right), \\
s^*_{ijc} &= z_T^{ijc} - \big[(1-\alpha^*_{ijc})z_A^{ijc} + \alpha^*_{ijc}z_B^{ijc}\big],
\end{align}
where $\Pi_{[0,1]}$ denotes projection onto $[0,1]$.  

This formulation interprets $\alpha^*$ as the projection of $z_T$ onto the line spanned by $(z_A,z_B)$, while $s^*$ captures the orthogonal residual.

\begin{figure}
    \section{Blind Predictor Architecture}
    \label{blind-pred-arch}
    \centering
    \includegraphics[width=1\linewidth]{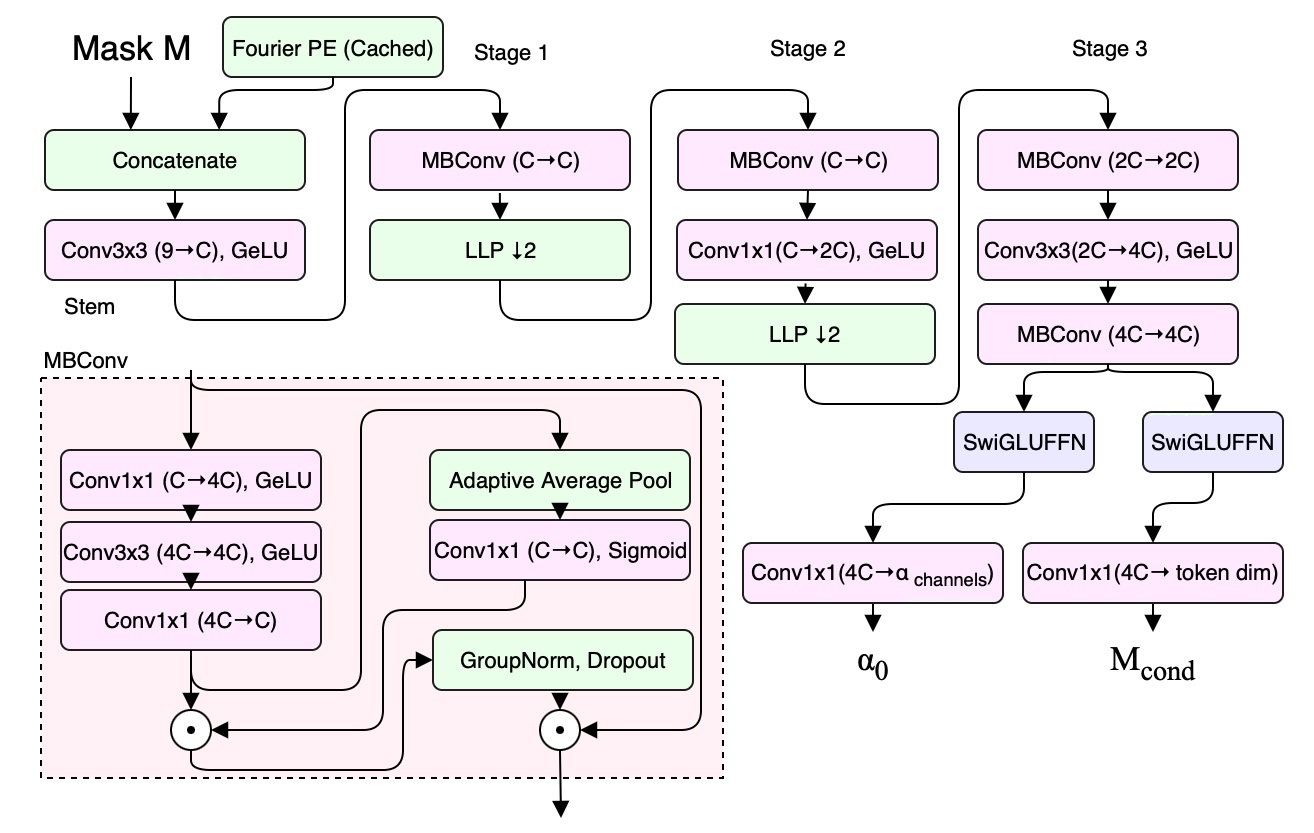}
    \caption{A lightweight CNN maps the input pixel mask (augmented with Fourier features) into latent-resolution, per-channel soft masks. The stem is a $3\times 3$ conv with GELU, followed by three stages: Stage 1: MBConv block with depth-wise squeeze–excite, followed by a learnable low-pass filter and $2\times$ downsampling; Stage 2: MBConv → pointwise expansion → GELU → learnable low-pass, giving another $2\times$ downsample; Stage 3: MBConv → strided conv for $8\times$ total reduction → MBConv (extra receptive field). Final shared features branch into two FFNGlU heads: (i) an $\alpha$ head predicting per-channel blending masks with bounded activation, and (ii) a token head producing spatial embeddings for cross-attention in DecFormer. The diagram expands the MBConv block (pointwise–depthwise–pointwise with SE and normalization); the learnable low-pass filters are depth-wise convolutions initialized as binomial blur kernels.}
\end{figure}

\begin{figure}[!h]
    \section{DecFormer Extended Architecture}
    \label{deltaformer-extended-arch}
    \centering
    \includegraphics[width=\linewidth]{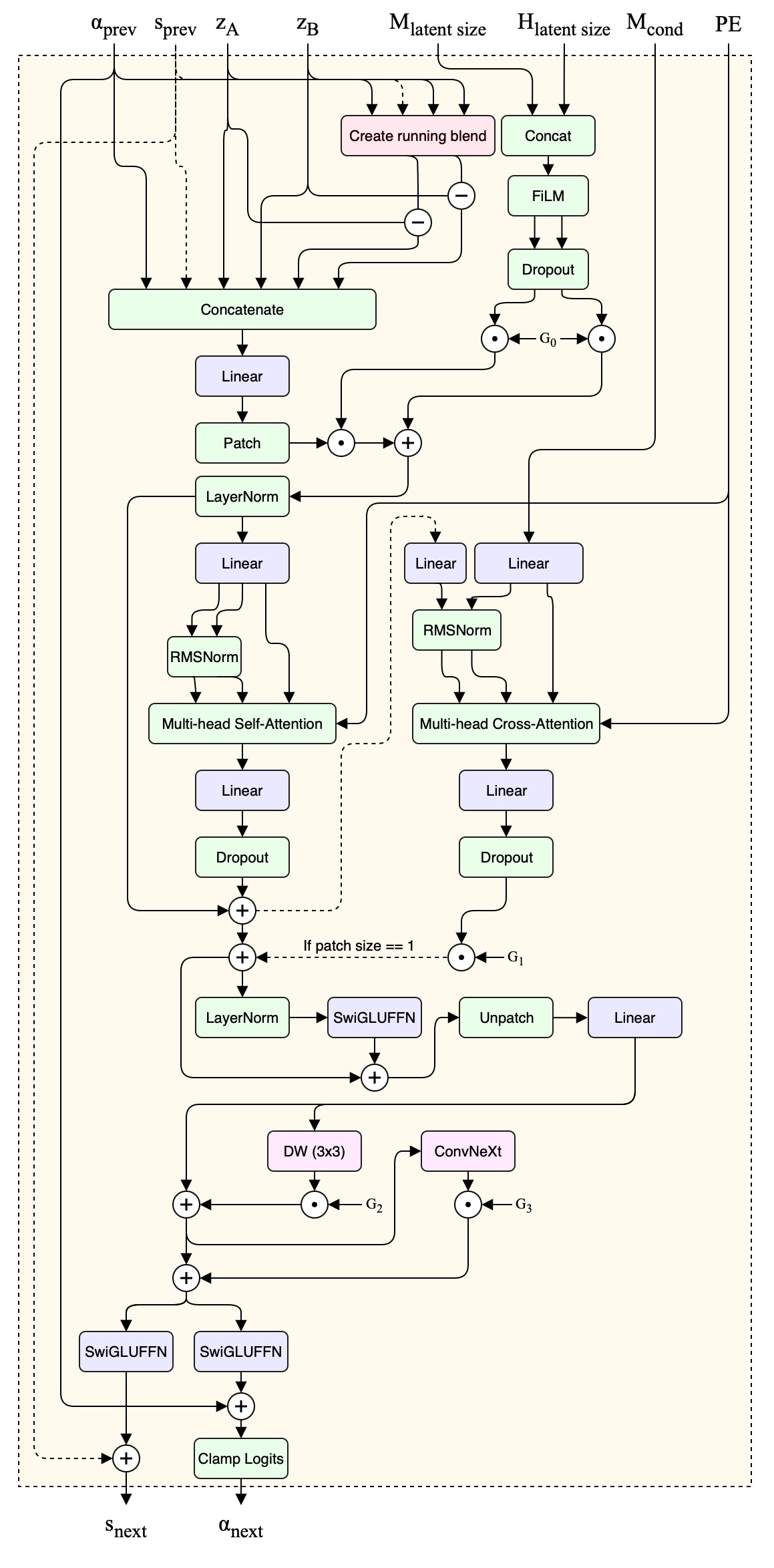}
    \caption{Extended DecFormer architecture diagram. The figure highlights more precisely the internal composition of each block, including the location and type of normalization layers, as well as the flow of intermediate projections and residual connections.}
    \label{fig:internals}
\end{figure}

\begin{figure}[!h]
    \section{Gamma Correction Proof-of-Concept Qualitative Results}
    \label{non-compositing-img}
    \centering
    \includegraphics[width=0.75\linewidth]{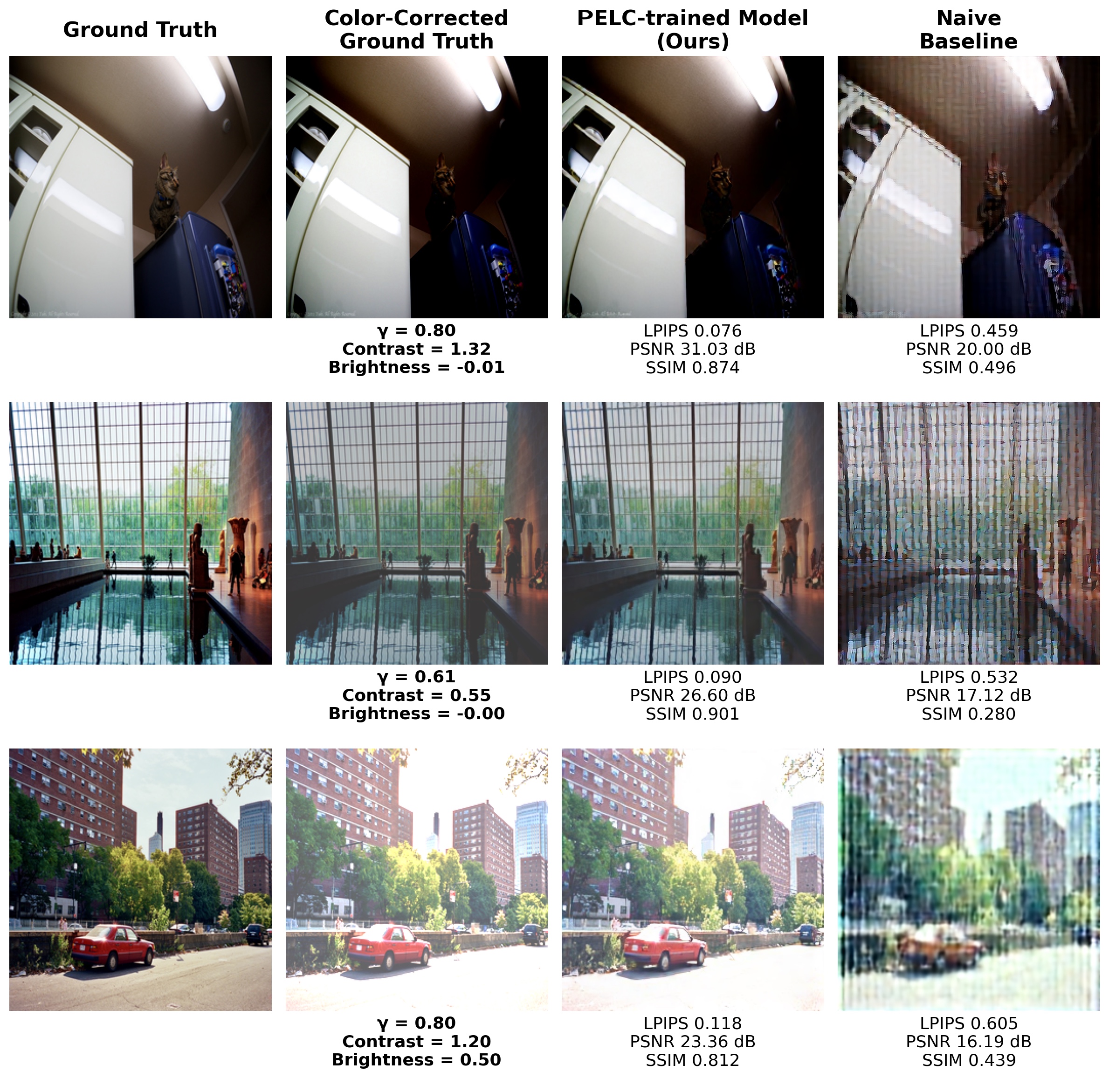}
    \caption{A qualitative example shows how the naive method fails catastrophically with severe artifacts when faced with more aggressive color correction, while our model's output is nearly indistinguishable from the ground truth. This illustrates that VAE latents are not pixel-like and require a principled framework like PELC.}
    \label{fig:gamma_images_vertical}
\end{figure}

\clearpage

\section{Extended DecFormer Metrics Tables}
\label{extended-metrics}
\begin{table}[h!]
\centering
\caption{Complete method comparison at 1024px resolution (mean ± 95\% CI, n=50). DecFormer variants compared against all heuristic baselines.}
\label{tab:full_methods_1024}
\resizebox{0.6\textwidth}{!}{%
\begin{tabular}{llcccc}
\toprule
\textbf{Mask Type} & \textbf{Method} & \textbf{SSIM} $\uparrow$ & \textbf{PSNR (dB)} $\uparrow$ & \textbf{LPIPS} $\downarrow$ & \textbf{Halo L1} $\downarrow$ \\
\midrule
\multirow{6}{*}{Soft ($\sigma$=21)} & DecFormer & 0.985$_{\pm.003}$& \textbf{41.3}$_{\pm.8}$ & \textbf{0.027}$_{\pm.005}$ & \textbf{0.018}$_{\pm.001}$ \\
 & DecFormer-Pretrain & \textbf{0.986}$_{\pm.002}$& 40.9$_{\pm.7}$ & 0.028$_{\pm.005}$ & \textbf{0.018}$_{\pm.001}$\\
 & Heuristic Area & 0.941$_{\pm.010}$ & 32.9$_{\pm1.1}$ & 0.088$_{\pm.016}$ & 0.050$_{\pm.005}$ \\
 & Heuristic Bilinear & 0.941$_{\pm.010}$ & 32.9$_{\pm1.1}$ & 0.088$_{\pm.016}$ & 0.050$_{\pm.005}$ \\
 & Heuristic Nearest & 0.940$_{\pm.010}$ & 32.4$_{\pm1.0}$ & 0.089$_{\pm.016}$ & 0.054$_{\pm.004}$ \\
\cmidrule(l){2-6}
\multirow{6}{*}{Binary} & DecFormer & \textbf{0.964}$_{\pm.017}$ & \textbf{35.7}$_{\pm1.5}$ & \textbf{0.045}$_{\pm.018}$ & \textbf{0.060}$_{\pm.006}$ \\
 & DecFormer-Pretrain & 0.961$_{\pm.018}$ & 34.8$_{\pm1.5}$ & 0.058$_{\pm.022}$ & 0.068$_{\pm.005}$ \\
 & Heuristic Area & 0.915$_{\pm.025}$ & 29.2$_{\pm1.2}$ & 0.112$_{\pm.029}$ & 0.135$_{\pm.007}$ \\
 & Heuristic Bilinear & 0.913$_{\pm.025}$ & 28.4$_{\pm1.3}$ & 0.110$_{\pm.029}$ & 0.141$_{\pm.008}$ \\
 & Heuristic Nearest & 0.903$_{\pm.028}$ & 26.3$_{\pm1.2}$ & 0.115$_{\pm.030}$ & 0.183$_{\pm.010}$ \\
\cmidrule(l){2-6}
\multirow{6}{*}{Original} & DecFormer & \textbf{0.968}$_{\pm.016}$ & \textbf{38.6}$_{\pm1.5}$ & \textbf{0.049}$_{\pm.018}$ & \textbf{0.037}$_{\pm.005}$ \\
 & DecFormer-Pretrain & 0.965$_{\pm.017}$ & 37.9$_{\pm1.4}$ & 0.056$_{\pm.021}$ & 0.040$_{\pm.005}$ \\
 & Heuristic Area & 0.919$_{\pm.024}$ & 31.5$_{\pm1.3}$ & 0.104$_{\pm.028}$ & 0.078$_{\pm.006}$ \\
 & Heuristic Bilinear & 0.918$_{\pm.024}$ & 31.1$_{\pm1.4}$ & 0.104$_{\pm.028}$ & 0.080$_{\pm.007}$ \\
 & Heuristic Nearest & 0.907$_{\pm.027}$ & 28.9$_{\pm1.2}$ & 0.110$_{\pm.030}$ & 0.110$_{\pm.009}$ \\
\cmidrule(l){2-6}
\multirow{6}{*}{Thin} & DecFormer & \textbf{0.967}$_{\pm.014}$ & \textbf{34.7}$_{\pm1.5}$ & \textbf{0.045}$_{\pm.017}$ & \textbf{0.073}$_{\pm.005}$ \\
 & DecFormer-Pretrain & 0.960$_{\pm.016}$ & 33.4$_{\pm1.4}$ & 0.061$_{\pm.020}$ & 0.085$_{\pm.005}$ \\
 & Heuristic Area & 0.922$_{\pm.029}$ & 28.6$_{\pm1.2}$ & 0.112$_{\pm.032}$ & 0.167$_{\pm.008}$ \\
 & Heuristic Bilinear & 0.920$_{\pm.030}$ & 27.3$_{\pm1.2}$ & 0.111$_{\pm.031}$ & 0.174$_{\pm.009}$ \\
 & Heuristic Nearest & 0.908$_{\pm.034}$ & 25.6$_{\pm1.3}$ & 0.116$_{\pm.032}$ & 0.207$_{\pm.011}$ \\
\bottomrule
\end{tabular}
}
\end{table}

\begin{table}[h!]
\centering
\caption{DecFormer vs. Heuristic bilinear at 512px resolution (mean ± 95\% CI, n=50).}
\label{tab:results_512px}
\resizebox{0.6\textwidth}{!}{%
\begin{tabular}{llcccc}
\toprule
\textbf{Mask Type} & \textbf{Method} & \textbf{SSIM} $\uparrow$ & \textbf{PSNR (dB)} $\uparrow$ & \textbf{LPIPS} $\downarrow$ & \textbf{Halo L1} $\downarrow$ \\
\midrule
\multirow{2}{*}{Soft ($\sigma$=21)} & DecFormer & \textbf{0.957}$_{\pm.008}$ & \textbf{36.0}$_{\pm.9}$ & \textbf{0.051}$_{\pm.009}$ & \textbf{0.029}$_{\pm.003}$ \\
 & Heuristic Bilinear & 0.859$_{\pm.024}$ & 28.6$_{\pm1.0}$ & 0.149$_{\pm.021}$ & 0.072$_{\pm.007}$ \\
\cmidrule(l){2-6}
\multirow{2}{*}{Binary} & DecFormer & \textbf{0.924}$_{\pm.028}$ & \textbf{31.0}$_{\pm1.5}$ & \textbf{0.069}$_{\pm.022}$ & \textbf{0.087}$_{\pm.011}$ \\
 & Heuristic Bilinear & 0.848$_{\pm.037}$ & 25.2$_{\pm1.2}$ & 0.145$_{\pm.029}$ & 0.168$_{\pm.010}$ \\
\cmidrule(l){2-6}
\multirow{2}{*}{Original} & DecFormer & \textbf{0.930}$_{\pm.026}$ & \textbf{33.1}$_{\pm1.5}$ & \textbf{0.068}$_{\pm.021}$ & \textbf{0.064}$_{\pm.009}$ \\
 & Heuristic Bilinear & 0.853$_{\pm.036}$ & 27.1$_{\pm1.3}$ & 0.139$_{\pm.029}$ & 0.123$_{\pm.011}$ \\
\cmidrule(l){2-6}
\multirow{2}{*}{Thin} & DecFormer & \textbf{0.942}$_{\pm.018}$ & \textbf{30.7}$_{\pm1.1}$ & \textbf{0.063}$_{\pm.018}$ & \textbf{0.091}$_{\pm.007}$ \\
 & Heuristic Bilinear & 0.878$_{\pm.034}$ & 24.4$_{\pm1.0}$ & 0.139$_{\pm.028}$ & 0.193$_{\pm.013}$ \\

\bottomrule
\end{tabular}
}
\end{table}

\begin{table}[h!]
\centering
\caption{DecFormer vs. Heuristic bilinear at 256px resolution (mean ± 95\% CI, n=50)}
\label{tab:results_256px}
\resizebox{0.6\textwidth}{!}{%
\begin{tabular}{llcccc}
\toprule
\textbf{Mask Type} & \textbf{Method} & \textbf{SSIM} $\uparrow$ & \textbf{PSNR (dB)} $\uparrow$ & \textbf{LPIPS} $\downarrow$ & \textbf{Halo L1} $\downarrow$ \\
\midrule
\multirow{2}{*}{Soft ($\sigma$=21)} & DecFormer & \textbf{0.934}$_{\pm.007}$ & \textbf{33.0}$_{\pm.7}$ & \textbf{0.071}$_{\pm.007}$ & \textbf{0.034}$_{\pm.003}$ \\
 & Heuristic Bilinear & 0.804$_{\pm.019}$ & 26.0$_{\pm.7}$ & 0.204$_{\pm.018}$ & 0.082$_{\pm.006}$ \\
\cmidrule(l){2-6}
\multirow{2}{*}{Binary} & DecFormer & \textbf{0.892}$_{\pm.029}$ & \textbf{27.9}$_{\pm1.3}$ & \textbf{0.098}$_{\pm.026}$ & \textbf{0.097}$_{\pm.013}$ \\
 & Heuristic Bilinear & 0.808$_{\pm.037}$ & 23.0$_{\pm1.0}$ & 0.187$_{\pm.032}$ & 0.172$_{\pm.013}$ \\
\cmidrule(l){2-6}
\multirow{2}{*}{Original} & DecFormer & \textbf{0.902}$_{\pm.027}$ & \textbf{29.7}$_{\pm1.3}$ & \textbf{0.097}$_{\pm.026}$ & \textbf{0.077}$_{\pm.009}$ \\
 & Heuristic Bilinear & 0.812$_{\pm.037}$ & 24.3$_{\pm1.0}$ & 0.183$_{\pm.033}$ & 0.142$_{\pm.010}$ \\
\cmidrule(l){2-6}
\multirow{2}{*}{Thin} & DecFormer & \textbf{0.911}$_{\pm.017}$ & \textbf{27.4}$_{\pm1.0}$ & \textbf{0.092}$_{\pm.017}$ & \textbf{0.097}$_{\pm.006}$ \\
 & Heuristic Bilinear & 0.809$_{\pm.032}$ & 21.5$_{\pm.8}$ & 0.202$_{\pm.027}$ & 0.199$_{\pm.010}$ \\

\bottomrule
\end{tabular}
}
\end{table}

\FloatBarrier

\begin{figure*}
    \centering
    \section{Alpha and Shift Visualisation and Target Visualisations}
    \label{alphashiftinternals}
    \centering
    \includegraphics[width=\textwidth]{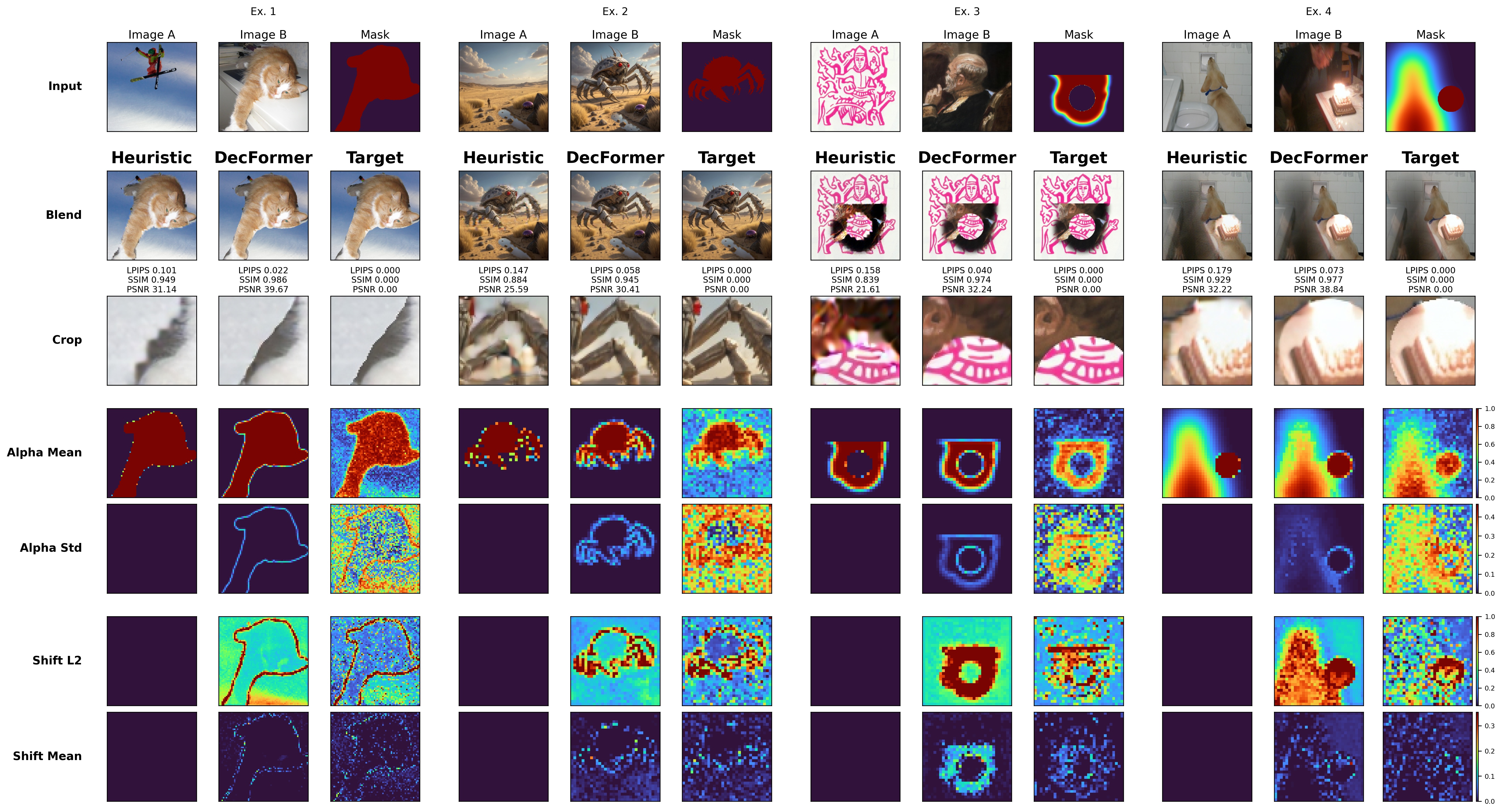}
    \caption{Qualitative comparison of DecFormer interpolation against a heuristic baseline and ground truth. For each method, we visualize the output alongside the corresponding $\alpha$ and shift predictions, compressed to 1-D profiles using multiple metrics. In the heuristic baseline, the naive mask collapses to a single scalar channel (zero variance), revealing its broadcast nature. The ground truth reference uses optimal $\alpha^*$, $s^*$ values (Appendix~\ref{app:projection}). Notably, the predicted shift and projected shift exhibits ring-like halos aligned with the mask boundaries, the latter of which was used to justify halo loss metrics and conditioning and ring radius.}
    \label{fig:decformer-internals}
\end{figure*}

\begin{figure*}
    \section{DecFormer Qualitative Extended}
    \label{hero-extended}
    \centering
    \includegraphics[width=1\textwidth]{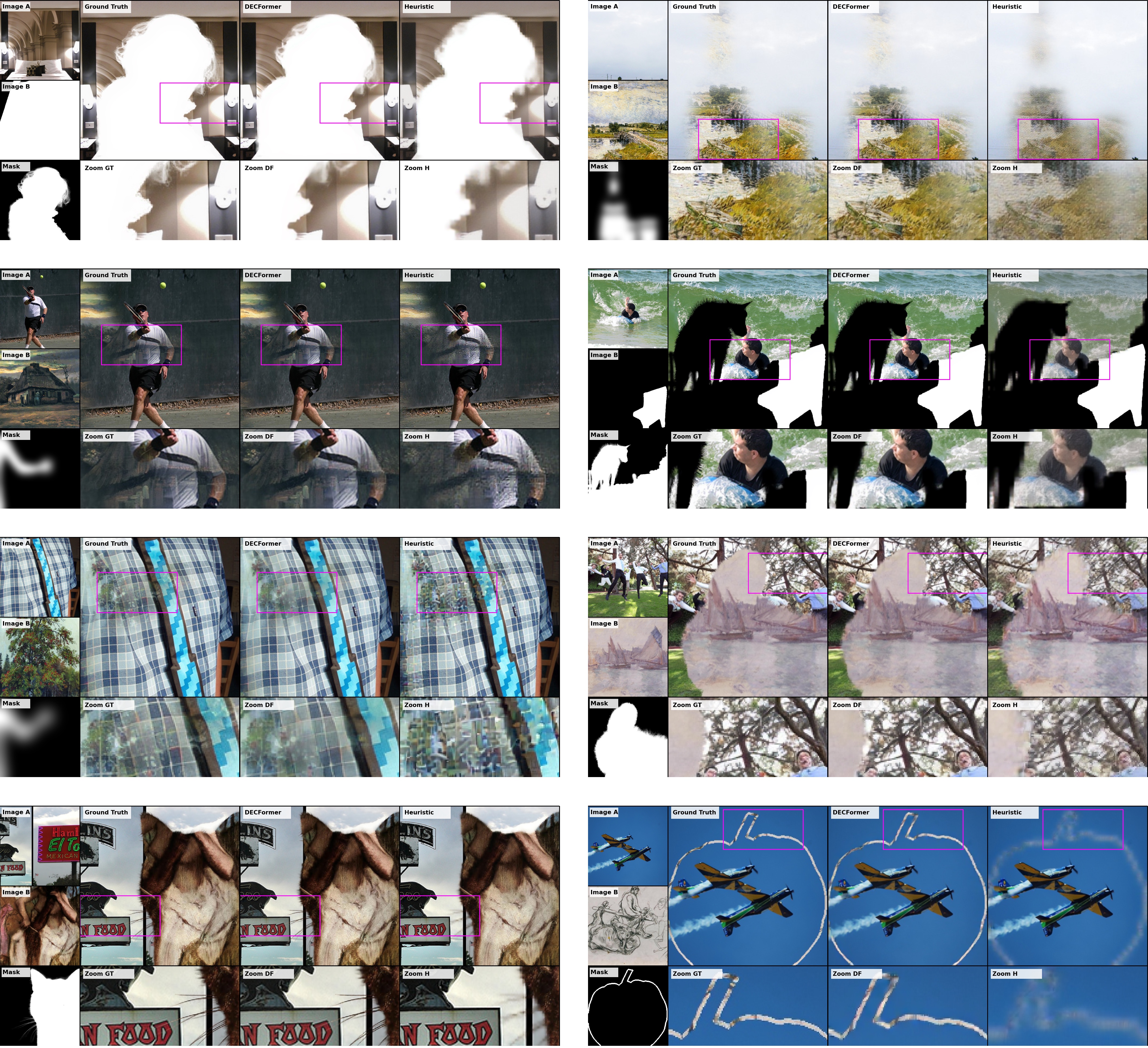}
    \caption{Further qualitative results illustrating the failure modes of heuristic interpolation and the improvements achieved by DecFormer (see Fig.~\ref{fig:hero}).}
\end{figure*}

\FloatBarrier
\clearpage

\section{Halo calculation}
\label{halo-calc}
Compute the 1-px edge set $e$ of $m$ (morphological XOR of 1-px dilate/erode). Convolve $e$ with a linear disk kernel of radius $R_{\text{px}}$ (We empirically find radius $R_{\text{px}}=8$ (approximately one VAE receptive field) provides good coverage) to obtain a two-sided, softly decaying ring $w^{\text{px}}\in[0,1]^{H\times W}$.
Anti-alias downsample $m$ to $m_\ell$, then reproduce the same ring construction at latent scale with radius $R_\ell=\max\!\big(1,\;\lfloor R_{\text{px}}/s\rceil\big)$, $s=\max(H/h,\,W/w)$, yielding $w^{\ell}\in[0,1]^{h\times w}$.

\subsection*{Staged training via local quadratic surrogate}
\label{sec:staging}

\textbf{Setup.}
Near a current iterate $(\alpha,s)$ we linearize the decoder $D$ and form a local Gauss--Newton surrogate for the decoded losses. This yields a quadratic objective
\[
\min_{\delta\alpha,\;\delta s}\;
\frac{1}{2}\!
\begin{bmatrix}\delta\alpha\\ \delta s\end{bmatrix}^{\!\top}
\underbrace{\begin{bmatrix} M & B \\ B^\top & N \end{bmatrix}}_{H_{\text{loc}}}
\begin{bmatrix}\delta\alpha\\ \delta s\end{bmatrix}
- \left\langle g,\begin{bmatrix}\delta\alpha\\ \delta s\end{bmatrix}\right\rangle,
\]
where $M\!\succ\!0$ and $N\!\succ\!0$ are the Gauss--Newton blocks for $\alpha$ and $s$, and $B$ captures their interaction (concentrated near mask boundaries).

\textbf{Block coordinate view.}
Training $\alpha$ first with $s{=}0$ and then $s$ with $\alpha$ frozen is equivalent to applying a block Gauss--Seidel step on the local system. The second stage solves the Schur--complement system
\[
S\,\delta s \;=\; r_s - B^\top M^{-1} r_\alpha,
\qquad
S \;=\; N - B^\top M^{-1} B,
\]
which can be interpreted as preconditioning the joint problem by $M$ along the blend axis.

\textbf{Conditioning implication (local surrogate).}
Let $\kappa(\cdot)$ denote the spectral condition number. For the preconditioned local system one obtains the bound
\[
\kappa_{\text{BCGD}} \;\le\; \kappa(M)\,\kappa(S),
\qquad
S = N - B^\top M^{-1} B \preceq N,
\]
so $\kappa_{\text{BCGD}}$ is no worse than using $N$ alone and improves as the coupling $B$ is explained by the $\alpha$-update. Empirically, we observe a reduced spectrum of $S$ (vs.\ $N$) concentrated at mask boundaries. \footnote{This statement is for the \emph{local quadratic surrogate} induced by a frozen decoder Jacobian and squared decoded losses; in practice we use LPIPS and halo weightings, for which Gauss--Newton is a standard approximation.}

\textbf{Practical schedule.}
Motivated by this decomposition, we \emph{stage} training:
(i) optimize $\alpha$ with $s$ gated off until validation stabilizes;
(ii) warm up the shift head over $2$k steps while reducing $\alpha$’s LR;
(iii) ramp in halo-weighted losses to focus $s$ on boundary residuals.
This preserves a clean early gradient signal for $\alpha$ and directs $s$ to the orthogonal residual where it is most needed.

\newpage
\FloatBarrier

\section{Receptive Field Analysis} \label{sec:RF analysis}

We calculate the receptive field according to formula:
$$
      r_0 = \sum_{l=1}^{L} \left((k_l-1)\prod_{i=1}^{l-1} s_i\right) + 1
$$
we modify this formula to find influence field
$$
      i_0 = \sum_{l=1}^{L} \left((k_l-1)\prod_{i=1}^{l-1} s_i \prod_{j=l-1}^{L} f_j \right) + \prod_{k=1}^{L} f_k
$$
where $f$ denotes the upscaling factor for each module, which will be 2 for linearly interpolating the hidden state, and 1 for other operations such as convolutions.

\begin{table}[h]
\centering
\caption{Encoder receptive field}
\resizebox{\columnwidth}{!}{%
\begin{tabular}{lrrr}
\toprule
\textbf{Layer} & \textbf{Effective Stride} & \textbf{Layer Sum} & \textbf{Cumulative field} \\
\midrule
Conv\_in & 1 & 2 & 3 \\
Down L0   & 1 & 10 & 13 \\
Down L1   & 2 & 20 & 33 \\
Down L2   & 4 & 40 & 73 \\
Down L3   & 8 & 64 & 137 \\
Middle     & 8 & 64 & 201 \\
Conv\_out & 8 & 16 & 217 \\
\bottomrule
\end{tabular}
}
\end{table}

\begin{table}[h]
\centering
\caption{Decoder influence field}
\resizebox{\columnwidth}{!}{%
\begin{tabular}{lrlrr}
\toprule
\textbf{Layer} & \textbf{Effective Stride}  & \textbf{Upscale Factor} & \textbf{Layer Sum} & \textbf{Cumulative field} \\
\midrule
Conv\_in & 1  &1& 2  & 3 \\
Middle     & 1  &1& 8  & 11 \\
Up L3     & 1  &2& 16 & 50 \\
Up L2     & 2  &2& 32 & 156 \\
Up L1     & 4  &2& 64 & 424 \\
Up L0     & 8  &1& 96 & 520 \\
Conv\_out & 8  &1& 16 & 536 \\
\bottomrule
\end{tabular}
}
\caption{Note that for upscaling layers, the order is reversed, from starting from L3 to L0}
\end{table}

\begin{table}[h]
\centering
\caption{Decoder receptive field}
\resizebox{\columnwidth}{!}{%
\begin{tabular}{lrrr}
\toprule
\textbf{Layer} & \textbf{Effective Stride} & \textbf{Layer Sum} & \textbf{Cumulative field} \\
\midrule
Conv\_in & 1 & 2 & 3 \\
Middle   & 1 & 8 & 11 \\
Up L3    & 1 & 13 & 24 \\
Up L2    & 1/2 & 6.5 & 30.5 \\
Up L1    & 1/4 & 3.25 & 33.75 \\
Up L0    & 1/8 & 1.5 & 35.25 \\
Conv\_out & 1/8 & 0.25 & 35.5 \\
\bottomrule
\end{tabular}
}
\end{table}

\section{Dual-Sigma Noise for Mask-Aware Inpainting LoRA}
\label{lora}
Standard diffusion applies a single global noise level $\sigma$ to all latent
locations,
\[
z_\sigma = (1-\sigma)z_0 + \sigma\,\epsilon,
\qquad
\epsilon\sim\mathcal{N}(0,I),
\]
yielding a spatially uniform SNR. This provides no mechanism to distinguish pixels that are context from the inpainting region. More critically, in few-step samplers and velocity-based flows, the first update step largely fixes the denoising trajectory. Under uniform corruption, the model begins this first step from a state in which the true context is fully noised, forcing a blind prediction. The resulting early commitment often drives the masked and unmasked regions toward different modes, producing visible seams or the appearance of two incompatible images interpolated together.

To avoid this failure mode, we impose a SNR contrast between the two regions. The surrounding context is assigned lower noise (higher SNR), while the masked region receives higher noise (lower SNR). This ensures that the model sees an accurate representation of the context at the first denoising step, providing a reliable conditioning signal before any irreversible trajectory decisions are made.
\paragraph{Dual-sigma construction}
Let $m\!\in\![0,1]^{H\times W}$ denote the latent-resolution mask. For each
sample we draw $u\!\sim\!\mathcal{U}(0,1)$ and construct
\[
\sigma_{\text{in}} = g(u),
\qquad
\sigma_{\text{out}} = g(\lambda u),
\]
where $\lambda=0.75 < 1$ is a fixed scalar controlling the
maximum noise level applied to the context region.

and thus a corresponding SNR contrast
\[
\mathrm{SNR}_{\text{out}}
= \frac{(1-\sigma_{\text{out}})^2}{\sigma_{\text{out}}^2}
\;\gg\;
\frac{(1-\sigma_{\text{in}})^2}{\sigma_{\text{in}}^2}
= \mathrm{SNR}_{\text{in}}.
\]

\paragraph{Regionwise noising}
Per-region noisy latents are constructed as
\[
z_{\sigma} = (1-\sigma)z_0 + \sigma\,\epsilon,
\]
and the composite latent is
\[
z
=
m \odot z_{\sigma_{\text{in}}}
+
(1-m) \odot z_{\sigma_{\text{out}}}.
\]

This yields a piecewise noise field
\[
\sigma(i,j)
=
m(i,j)\sigma_{\text{in}}
+
(1 - m(i,j))\sigma_{\text{out}},
\]
in contrast to the spatially uniform $\sigma$ used in standard diffusion. The denoiser naturally relies on the preserved context to guide reconstruction
of the masked region, producing an inpainting-aware model without modifying architecture. This modification is not significantly out of distribution, and is easy to learn in a low rank manner. We train a 16 rank LoRA on this formulation.
\end{document}